\documentclass[10pt,twocolumn,letterpaper]{article}

\usepackage{cvpr}
\usepackage{times}
\usepackage{epsfig}
\usepackage{graphicx}
\usepackage{amsmath}
\usepackage{amssymb}
\usepackage{epstopdf}
\usepackage{lipsum}
\usepackage{subfig}
\usepackage{enumitem}

\usepackage[pagebackref=true,breaklinks=true,letterpaper=true,colorlinks,bookmarks=false]{hyperref}

 \cvprfinalcopy 

\def\cvprPaperID{1814} 

\ifcvprfinal\fi
\begin{document}

\title{Mirror, mirror on the wall, tell me, is the error small?}

\author{Heng  Yang\\
Queen Mary University of London\\
{\tt\small heng.yang@qmul.ac.uk}
\and
Ioannis Patras\\
Queen Mary University of London\\
{\tt\small i.patras@qmul.ac.uk}
}

\maketitle

\begin{abstract}
Do object part localization methods produce bilaterally symmetric results on mirror images? Surprisingly not, even though state of the art methods augment the training set with mirrored images. In this paper we take a closer look into this issue. We first introduce the concept of mirrorability as the ability of a model to produce symmetric results in mirrored images and introduce a corresponding measure, namely the \textit{mirror error} that is defined as the difference between the detection result on an image  and the mirror of the detection result on its mirror image. We evaluate the mirrorability of several state of the art algorithms in two of the most intensively studied problems, namely human pose estimation and face alignment. Our experiments lead to several interesting findings: 1) Surprisingly, most of state of the art methods struggle to preserve the mirror symmetry, despite the fact that they do have very similar overall performance on the original and mirror images; 2) the low mirrorability is not caused by training or testing sample bias - all algorithms are trained on both the original images and their mirrored versions; 3) the mirror error is strongly correlated to the localization/alignment error (with correlation coefficients around 0.7). Since the mirror error is calculated without knowledge of the ground truth, we show two interesting applications - in the first it is used to guide the selection of difficult samples and in the second to give feedback in a popular Cascaded Pose Regression method for face alignment. 

\end{abstract}

\section{Introduction}
\begin{figure}

\subfloat[Mirror error 0.2.]{ \includegraphics[trim =3.5cm 2.5cm 4.5cm 0.0cm, clip = true,width=0.12\textwidth, height=0.2\textwidth]{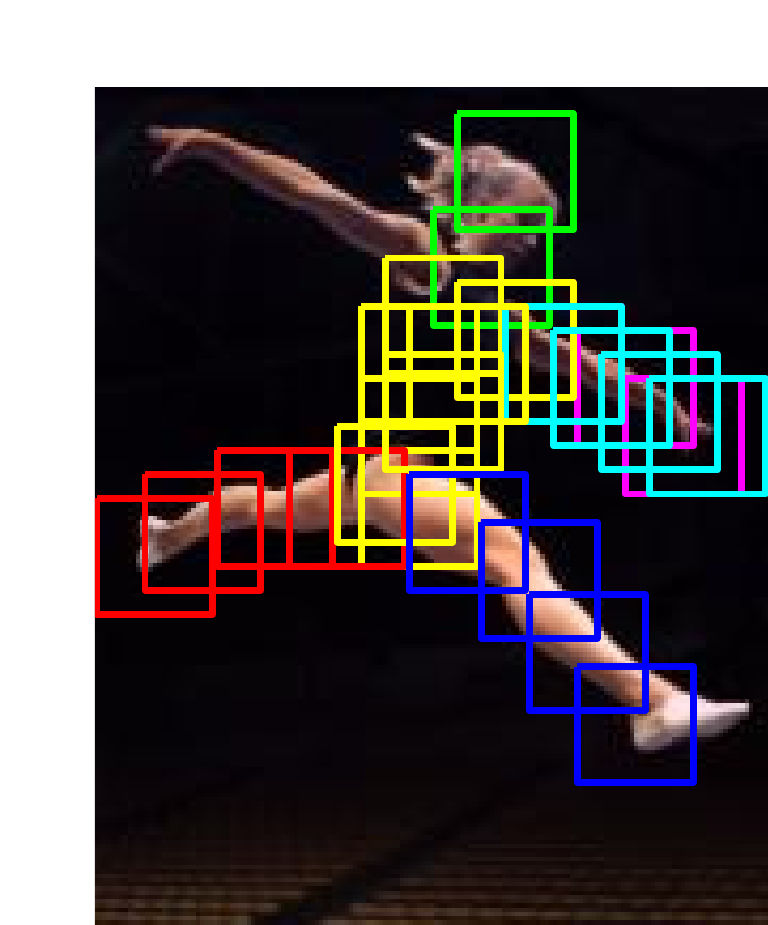}
\includegraphics[trim =4.5cm 2.5cm 3.5cm 0.0cm, clip = true,width=0.12\textwidth, height=0.2\textwidth]{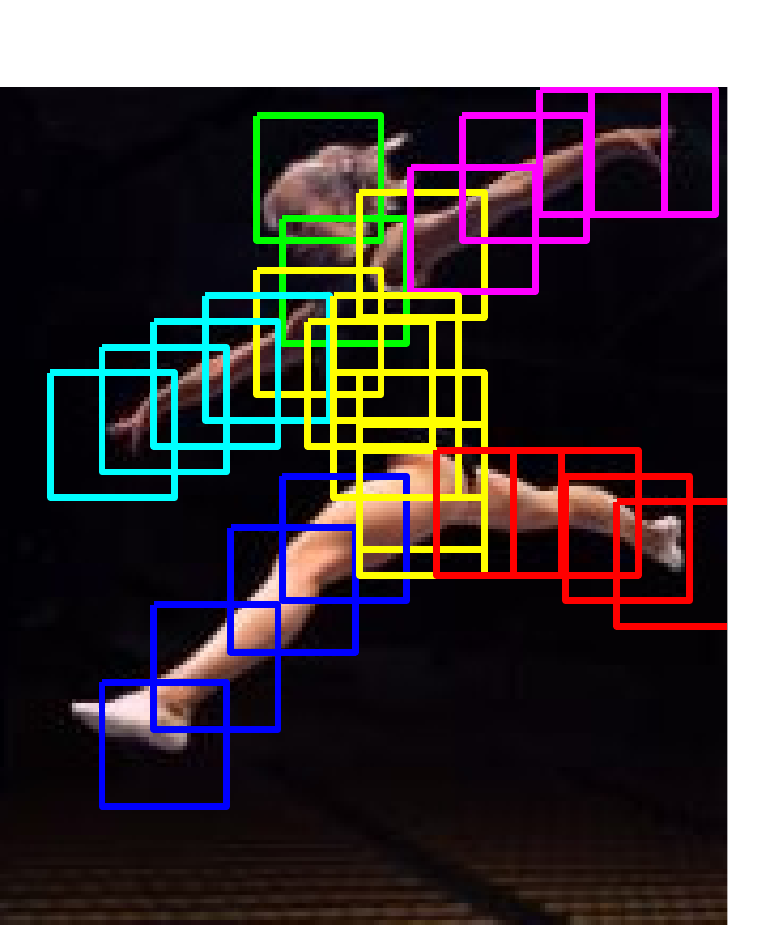}}
\subfloat[Mirror error 0.02.]{
\includegraphics[trim =4cm 1cm 5cm 1.0cm, clip = true,width=0.12\textwidth, height=0.18\textwidth]{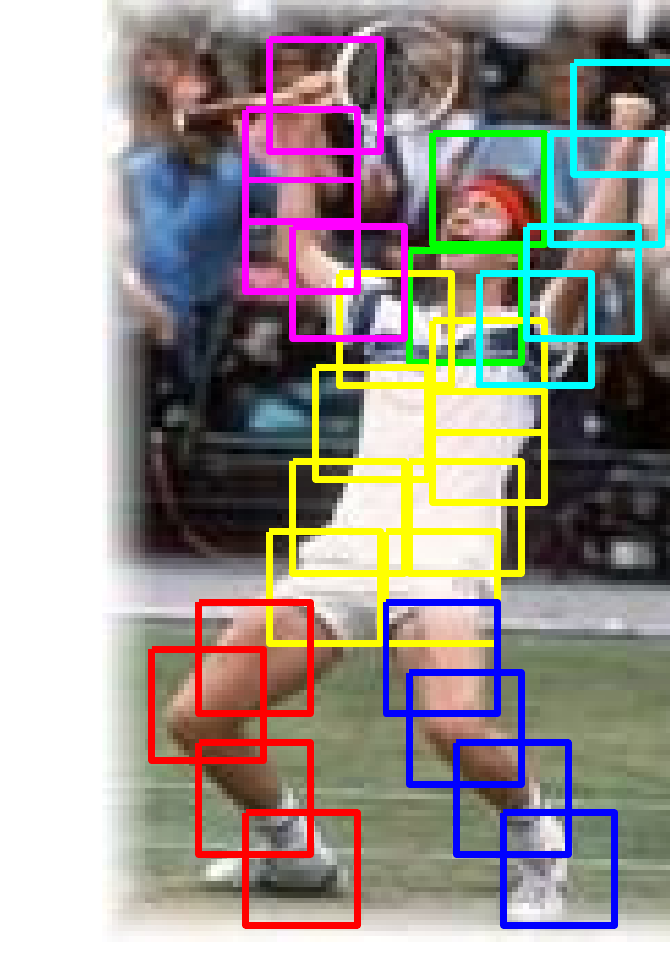}
\includegraphics[trim =5cm 1cm 4cm 1.0cm, clip = true,width=0.12\textwidth, height=0.18\textwidth]{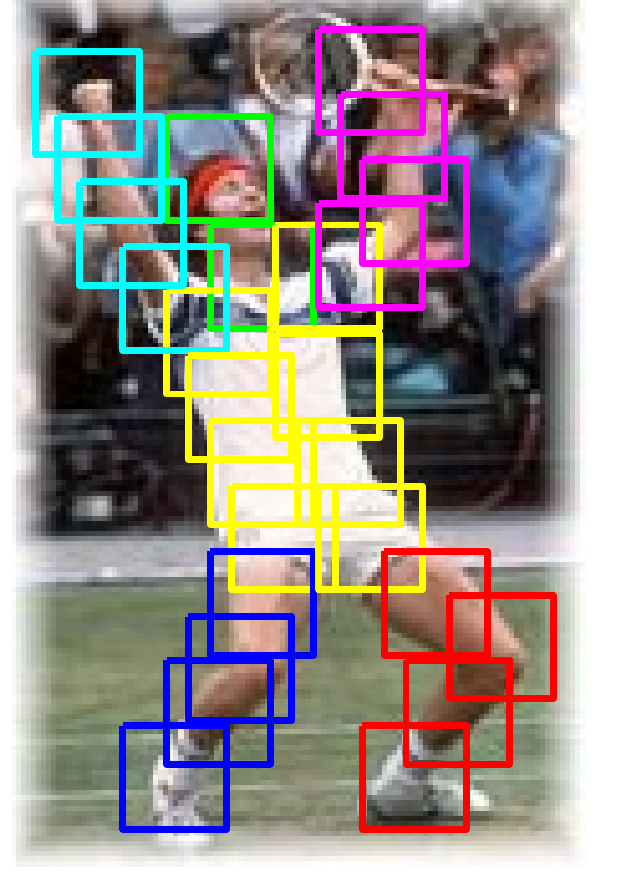}}

\subfloat[Mirror error 0.6.]{\includegraphics[trim =5cm 5.5cm 7cm 2.4cm, clip = true,width=0.12\textwidth, height=0.13\textwidth]{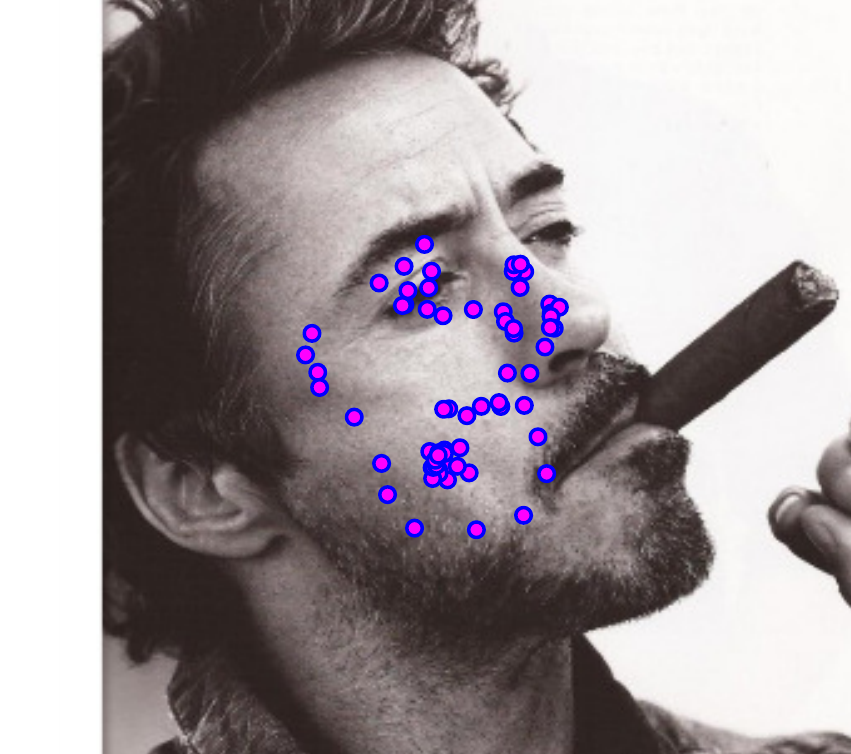}
\includegraphics[trim =7.6cm 5.5cm 5cm 2.4cm, clip = true,width=0.12\textwidth, height=0.13\textwidth]{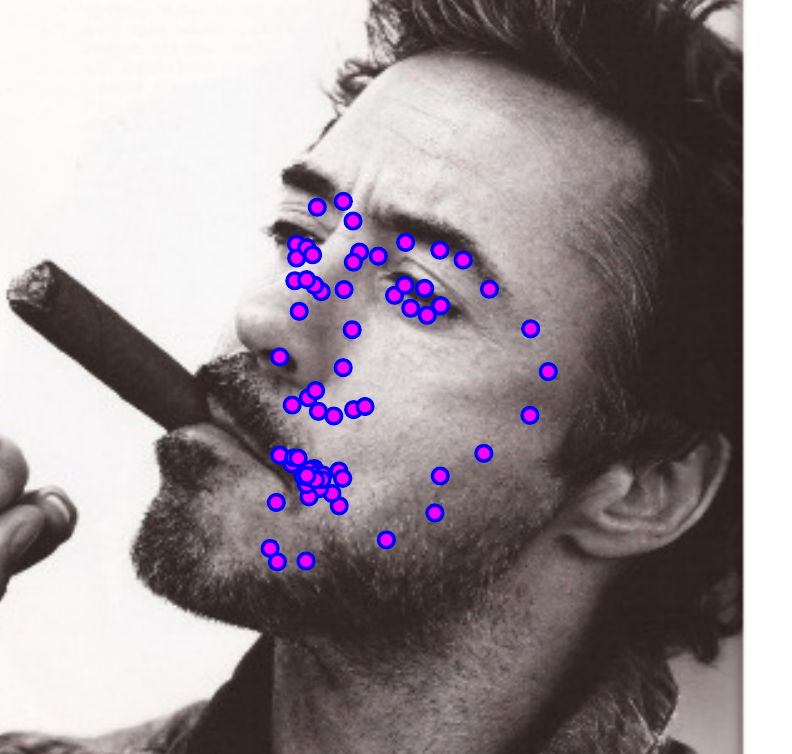}}
\subfloat[Mirror error 0.02.]{\includegraphics[trim =5cm 2.5cm 6.6cm 4.4cm, clip = true,width=0.12\textwidth, height=0.13\textwidth]{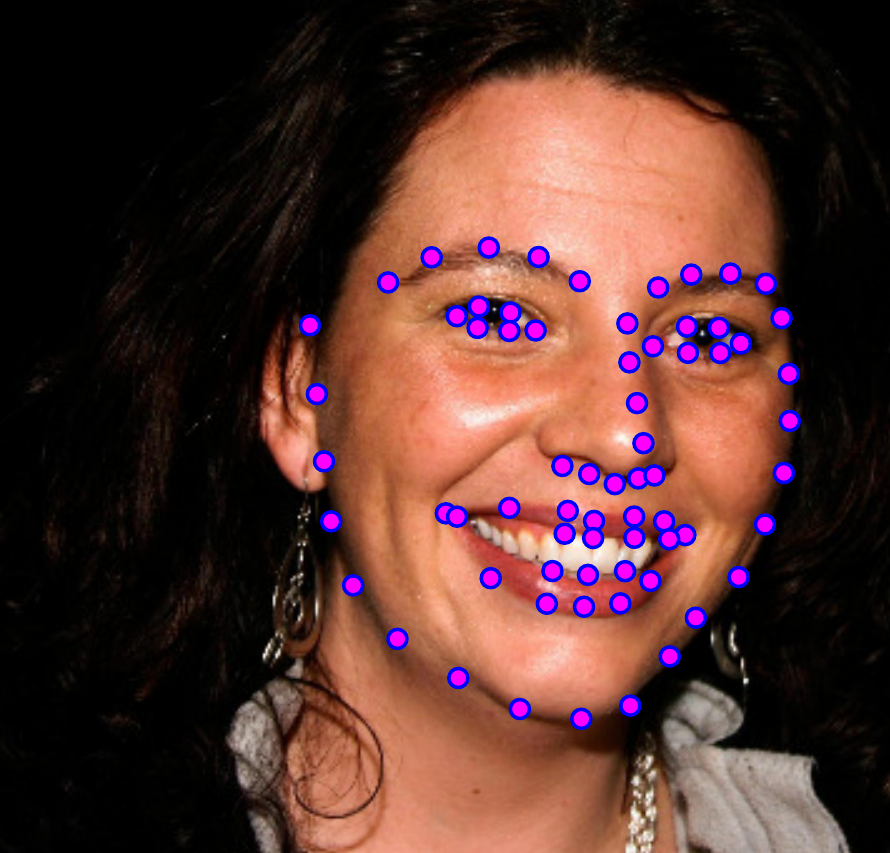}
\includegraphics[trim =6.6cm 2.5cm 5cm 4.4cm, clip = true,width=0.12\textwidth, height=0.13\textwidth]{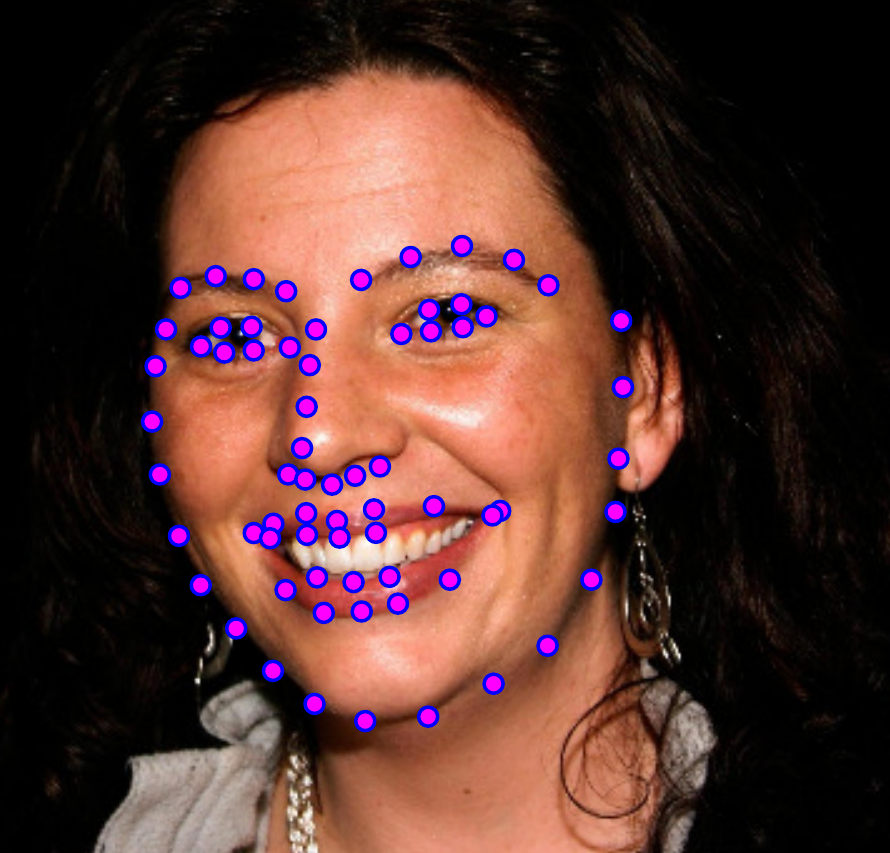}}

\caption{Example pairs of localization results on original (left) and mirror (right) images. First row: Human Pose Estimation \cite{yang2013articulated},  second row: Face Alignment by RCPR \cite{burgos2013robust}. The first column (a and c) shows large mirror error and the second (b and d) small mirror error. Can we evaluate the performance without knowing the ground truth?}
\label{fig::examples}
\end{figure}

The evolution of mirror (bilateral) symmetry has profoundly impacted animal evolution \cite{finnerty2005did}. As a consequence, the overwhelming majority of modern animals ($>$99\%), including humans, exhibit mirror symmetry. As shown in Fig.~\ref{fig::examples}, the mirror of an image depicting such objects shows a meaningful version of the same objects. Taking face images as a concrete example, a mirrored version of a face image is perceived as the same face.
In recent year, object (parts) localization has made significant progress and several methods have reported close-to-human performance. This includes localization of objects in images (e.g. pedestrian or face detection) or fine-grained localization of object parts (e.g. face parts localization, body parts localization, bird parts localization). Most of those methods augment the training set by mirroring the positive training samples. However, are these models able to give symmetric results on a mirror image during testing? 

In order to answer this question we first introduce the concept of mirrorability, i.e., the ability of an algorithm to give on a mirror image bilaterally symmetric results, and a quantitative measure called the mirror error. The latter is defined as the difference between the detection result on an image and the mirror of detection result on its mirror image. We evaluate the mirrorability of several state of the art algorithms in two representative problems (face alignment and human pose estimation) on several datasets. One would expect that a model that has been trained on a dataset augmented with mirror images to give similar results on an image and its mirrored version. However, as can be seen in Fig.~\ref{fig::examples} first column, several state of the art methods in their corresponding problems sometimes struggle to give symmetric results in the mirror images. And for some samples the mirror error is quite large. By looking at the mirrorability of different approaches in human pose estimation and face alignment, we arrive at three interesting findings. First, most of the models struggle to preserve the mirrorability - the mirror error is present and sometimes significant; Second, the low mirrorability is not likely to be caused by training or testing sample bias - the training sets are augmented with mirrored images; Third, the mirror error of the samples is highly correlated with the corresponding ground truth error. 

This last finding is significant since one of the \textit{nice} properties of the proposed mirror error is that it is calculated 'blindly', i.e., without using the ground truth. We rely on this property in order to show two examples of how it could be used in practice. In the first one the mirror error is used as a guide for difficult samples selection in unlabelled data and in the second one it is used to provide feedback on a cascaded pose regression method for face alignment. In the former application, the samples selected based on the mirror error have shown high consistency across different methods and high consistency with the difficult samples selected based on the ground truth alignment error. In the latter application, the feedback mechanism is used in a multiple initializations scheme in order to detect failures - this leads to large improvements and state of the art results in face alignment.

To summarize, in this paper we make the following contributions:
\begin{itemize}[topsep=0pt,itemsep=-1ex,partopsep=1ex,parsep=1ex]
\item To the best of our knowledge, we are the first to look into the mirror symmetric performance of object part localization models. 
\item We introduce the concept of mirrorability and show how the corresponding measure, called mirror error, that we propose can be used in evaluating general object part localization methods. 
\item We evaluate the mirrorability of several algorithms in two domains (i.e. face alignment and body part localization) and report several interesting findings on the mirrorability.
\item We show two applications of the mirrorability in the domain of face alignment. 
\end{itemize}


\section{Mirrorability in Object Part Localization}
\subsection{Mirrorability concepts and definitions}

We define mirrorability as the ability of a model/algorithm to preserve the mirror symmetry when applied on an image and its mirror image. 
In order to quantify it we introduce a measure called mirror error that is defined as the difference between a detection result on an image and the mirror of the result on its mirror image. Specifically, let us denote the shape of an object, for example a human or a face, by a set of $K$ points, $\mathrm{X} = \{\mathrm{x}_k\}_{k=1}^K$, where $\mathrm{x}_k$ are the coordinates of the $k$-th point/part. The detection result on the original image is denoted by $\mathrm{^qX} = \{\mathrm{^qx}_k\}_{k=1}^K$ and the detection result on the mirror image is denoted by $\mathrm{^pX} = \{\mathrm{^px}_k\}_{k=1}^K$. The mirror transformation of $\mathrm{^pX}$ to the original image is denoted by $\mathrm{^{p\rightarrow q} X} = \{\mathrm{^{p\rightarrow q}x}_k\}_{k=1}^K$, where $\mathrm{ ^{p\rightarrow q}x}_k$ denotes the mirror result of the $k$-th part on the original image. Generally, a different index $k'$ is used on the mirror image (e.g. a left eye in an image becomes a right eye in the mirror image). Therefore, the transformation consists of image coordinates transform and the part index mirror transform ($k'\rightarrow k$). The image coordinate transform is applied on the horizontal coordinate, that is ${^\mathrm{p}x_k} = \mathrm{w}_{I} - {^\mathrm{q}x_k}$, where $\mathrm{w}_{I}$ is the width of the image $I$ and $^\mathrm{p}x_k$ is the x coordinate of the $k$ point in the mirror image. The index re-assignment is based on the the mirror symmetric structure of a specific object, with an one-to-one mapping list  where, for example, the left eye index is mapped to the right eye index. Formally, the mirror error of the $k$ landmark (body joint or facial point) is defined as $||\mathrm{^qx}_k-\mathrm{^{p\rightarrow q}x }_k||$, and the sample-wise mirror error as:


\begin{equation}
e_{\text{m}} = \frac{1}{K}\sum_{k=1}^K||\mathrm{^qx}_k-\mathrm{^{p\rightarrow q}x }_k||
\label{eq::mirrorerror}
\end{equation}

The mirror error that is defined in the above equation has the following properties: First, a high mirror error reflects low mirrorability and vice visa; Second, it is symmetric, i.e., given a pair of mirror images it makes no difference which is considered to be the original; Third, and importantly, calculating the mirror error does not require ground truth information. 



In a similar way we calculate the ground truth localization error $\mathrm{^q}e_{\text{a}}$ as the difference between the detected locations and the ground truth locations of the facial landmarks or the human body joints. In order to be consistent and distinguish it from the mirror error we call it the alignment error. Formally,
\begin{equation}
\mathrm{^q}e_{\text{a}} = \frac{1}{K}\sum_{k=1}^K||\mathrm{^qx}_k-\mathrm{^{gt}x }_k||
\label{eq::alignmenterror}
\end{equation}
where $\mathrm{^{gt}x }_k$ is the ground truth location of the $k$-th point. In a similar way, we define the alignment error $\mathrm{^p}e_{\text{a}}$ on the mirror image of the test sample. For simplicity in what follows when we use the term of alignment error $e_a$, we mean the alignment error in the original image. 

Both Eq.~\ref{eq::mirrorerror} and Eq.~\ref{eq::alignmenterror} are absolute errors. In order to keep our analysis invariant to the size of the object in each image, we normalize them by the object size, i.e. $s$, the size of the body or the face. The size of the human body and the face are calculated in different ways and they are depicted when we use them. 
%
%

\subsection{Human pose estimation}
\paragraph{Experiment setting}
In order to evaluate the mirroability of algorithms for human pose estimation, we focus on two representative methods, namely the Flexible Mixtures of Parts (FMP) method by Yang and Ramanan \cite{yang2013articulated} and the Latent Tree Models (LTM) by Wang and Li \cite{fwang2013pose}. The FMP  is generally regarded as a benchmark method for human pose estimation and most of the recent methods are improved versions or variants of it. The one by Wang and Li \cite{fwang2013pose} introduced latent variables in tree model learning that led to improvements. Both of them have provided source code which we used in our evaluation. Since it is not our main focus to improve the performance in a specific domain, we use popular state of the art approaches and evaluate them on standard datasets. We use three widely used datasets, namely the Leeds Sport Dataset (LSP), the  Image Parse dataset \cite{ramanan2006learning} and the Buffy Stickmen dataset \cite{eichner20122d}. We use the default training/test split of the datasets. The number of test images on LSP, Parse and Buffy is 1000, 276 and 205 respectively. 
We trained both FMP and LTM models on LSP and only FMP model on Parse and Buffy. We emphasize that the training dataset is augmented with mirror images - this eliminates the training sample bias.

\noindent \textbf{Overall performance difference} We first compare the overall performance on the original test set and on the mirror set. We use the evaluation criterion proposed in \cite{yang2013articulated} and also recommended in \cite{andriluka14cvpr}, namely the Percentage of Correct Keypoints (PCK).  In order to calculate the PCK for each person a tightly-cropped bounding box is generated as the tightest box around the person in question that contains all of the ground truth keypoints. The size of the person is calculated as $s = \text{max}(h,w)$, where $h$ and $w$ are the hight and width of the bounding box. This is used to normalize the absolute mirror error in Eq.~\ref{eq::mirrorerror} and the alignment error in Eq.~\ref{eq::alignmenterror}.  The results on Buffy, Parse and LSP are shown in Table \ref{tab::buffypck}, Table \ref{tab::parsepck} and Table \ref{tab::lsppck} respectively. As can be seen, there is no significant overall difference between the detection results on the original images and on their mirror images. The maximum difference of different methods on different datasets is around 1\% while the average difference less than 1\%.

\begin{table}
\centering
\begin{tabular}{ccccccc}
\hline 
Points & Head & Shou & Elbo & Wri & Hip  & \textit{Avg} \\
Original      & 96.9 & 97.3 & 91.1 & 80.8 & 79.6&89.1 \\
Mirror       & 97.1 & 98.4 & 91.8  & 81.9 & 80.4&89.9 \\
\hline 
\end{tabular}
\caption{PCK of FMP \cite{yang2013articulated} on Buffy. A point is correct if the error is less than $0.2*\text{max}(h,w)$ }
\label{tab::buffypck}
\end{table}

\begin{table}
\centering
\small
\begin{tabular}{c@{\hskip 0.07in}c@{\hskip 0.07in}c@{\hskip 0.07in}c@{\hskip 0.07in}c@{\hskip 0.07in}c@{\hskip 0.07in}c@{\hskip 0.07in}c@{\hskip 0.07in}c}
\hline 
Points & Head & Shou & Elbo & Wris & Hip  & Knee & Ankle&\textit{Avg}\\
Original       & 90.0 & 85.6 & 68.3 & 47.3 & 77.3 & 75.6 & 67.3&73.1\\
Mirror       & 90.0 & 86.1 & 67.6 & 46.3 & 76.8 & 74.6 & 68.5&72.8  \\
\hline 
\end{tabular}
\caption{PCK of FMP \cite{yang2013articulated}  on Parse. A point is correct if the error is less than $0.1*\text{max}(h,w)$.}
\label{tab::parsepck}
\end{table}

\begin{table}
\centering
\small
\begin{tabular}{c@{\hskip 0.07in}c@{\hskip 0.07in}c@{\hskip 0.07in}c@{\hskip 0.07in}c@{\hskip 0.07in}c@{\hskip 0.07in}c@{\hskip 0.07in}c@{\hskip 0.07in}c}
\hline 
Points & Head & Shou & Elbo & Wris & Hip  & Knee & Ankle&\textit{Avg}\\
FMP Original&81.2 & 61.1 & 45.5&  33.4  &63.0&    55.6 & 49.5& 55.6\\
FMP Mirror &82.2  &61.0&  44.9&  33.8&  63.7    &56.1  &50.5& 56.0\\
LTM Original & 88.5&66.0&51.3&41.1&69.7&59.2&55.6 &61.6\\
LTM Mirror&88.7&65.8&51.4&40.7&70.2&58.0&55.0 &61.4\\

\hline 
\end{tabular}
\caption{PCK of FMP \cite{yang2013articulated} and LTM \cite{fwang2013pose} on LSP. A point is correct if the error is less than $0.1*\text{max}(h,w)$.}
\label{tab::lsppck}
\end{table}

%
\begin{figure}
\subfloat[Yang and Ramanan \cite{yang2013articulated}]{\includegraphics[trim =4cm 1cm 4cm 2cm, clip = true, width=0.25\textwidth,height=0.27\textwidth]{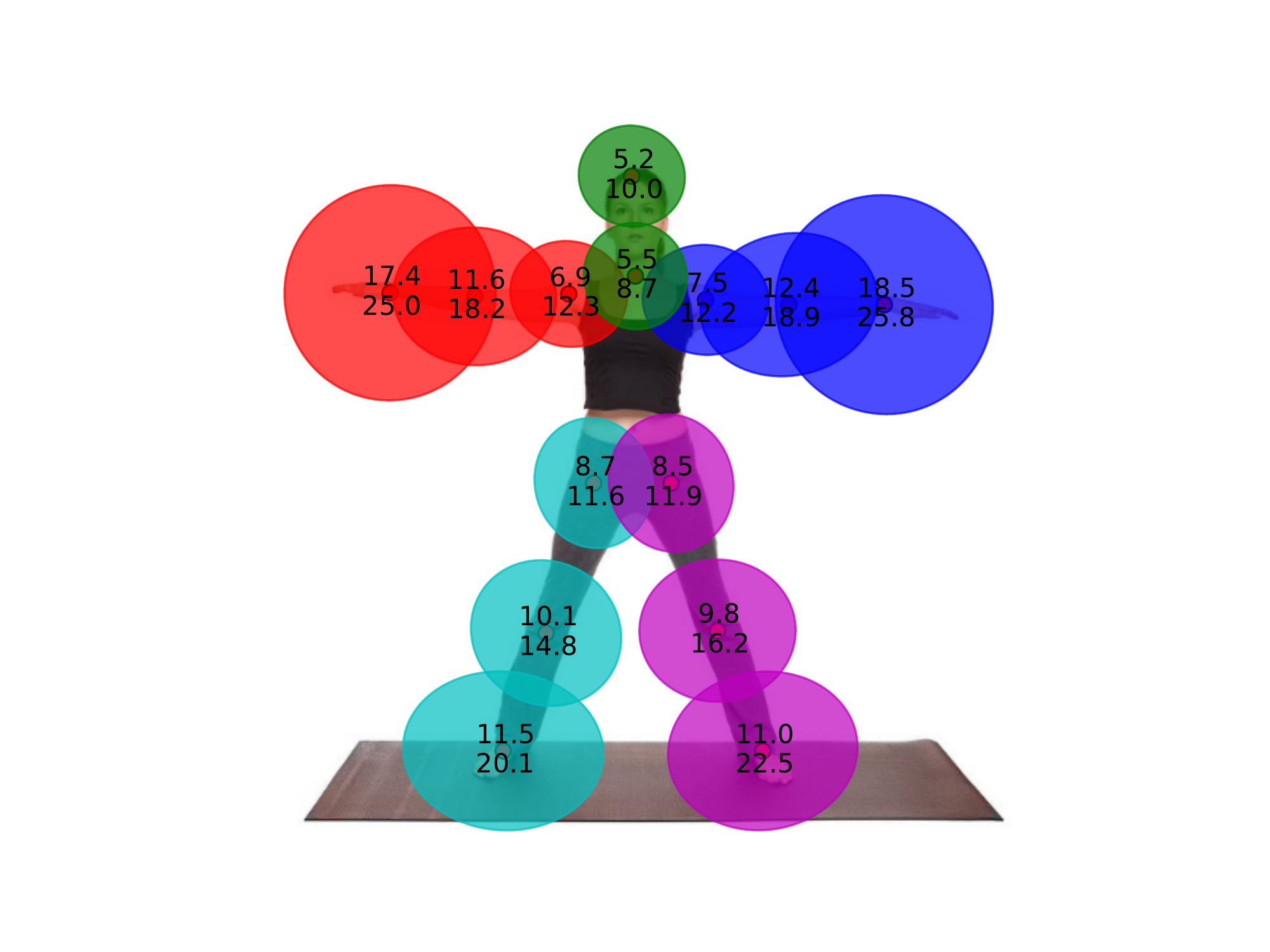}}
\subfloat[Wang and Li \cite{fwang2013pose}]{\includegraphics[trim =4cm 1cm 4cm 2cm, clip = true, width=0.25\textwidth,height=0.27\textwidth]{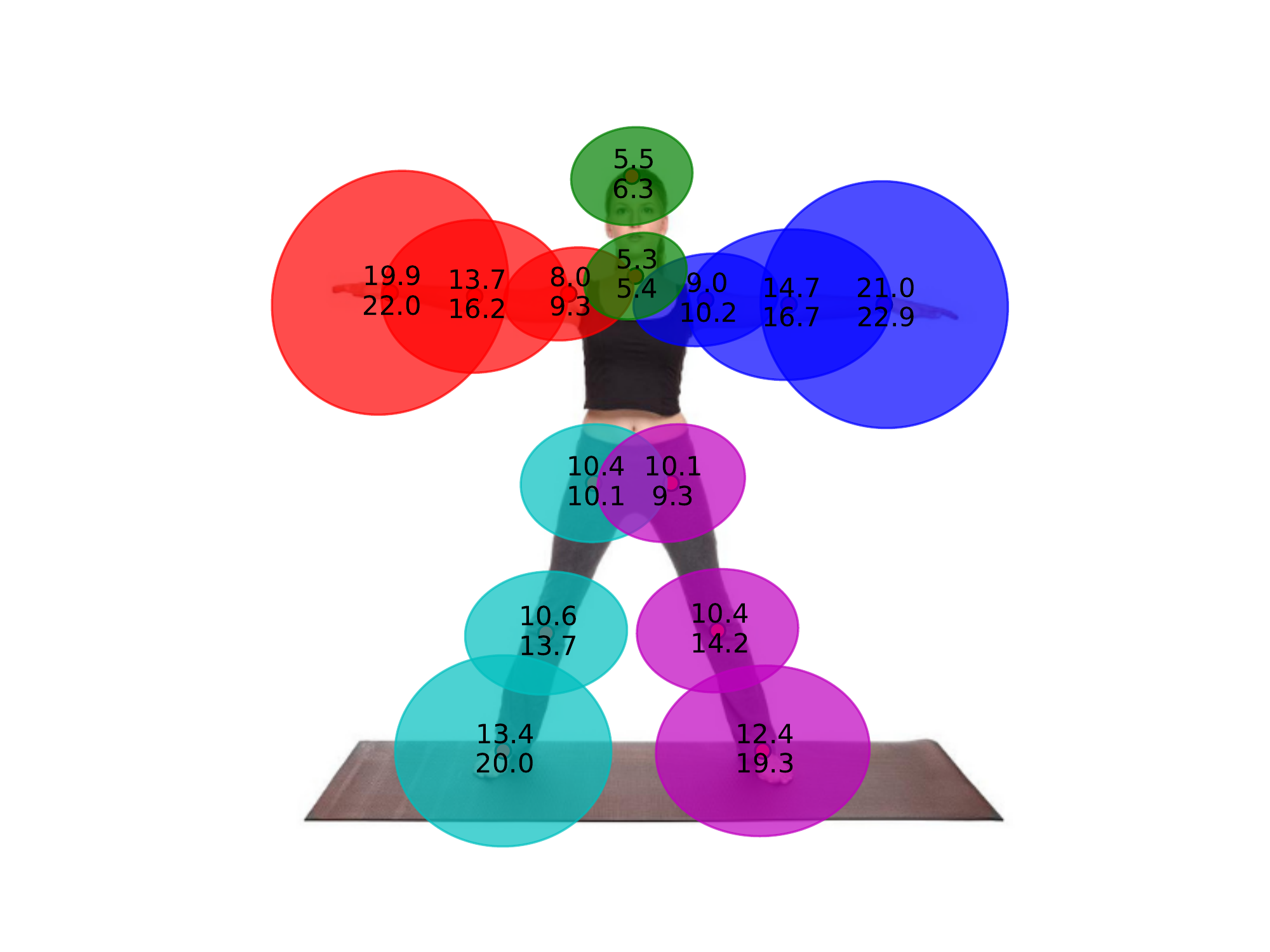}}
\caption{Visualization of mirror error (numbers on the upper) and alignment error (values on the lower) of body joints. The values are percentages of the body size.  The radius of each ellipse represents the value of one standard deviation of the mirror error on the corresponding body joint. }
\label{fig::visualizederror}
\end{figure}

\paragraph{Mirrorability}
The fact that the average performance on mirror images is similar to the average performance on the originals might be the root of the common belief that models produce more or less bilaterally symmetrical results. A closer inspection however reveals that this is not true. Let us first visualize the mirror error of individual body joints, i.e., $||\mathrm{^qx}_k-\mathrm{^{p\rightarrow q}x }_k||$ of both FMP and LTM on the LSP dataset. In Fig~\ref{fig::visualizederror} we plot the mirror error (normalized by the body size in the example image) of the 1000 test images on each individual joint.  As can be seen, there is a difference which in some cases it is quite large. For example on the elbows, feet and especially on the wrists ($\sim 18\% $ for FMP and $\sim 20\%$ for LTM).  This result directly challenges the perception that the models give mirror symmetrical results. We reiterate that this is despite the fact that the overall performance is similar in the original and the mirror images and despite the fact that we have augmented the training set with the mirror images. This leads us to the conclusion that the low mirrorability (i.e. large mirror error) is not the result of sample bias. 

\begin{figure}
\includegraphics[trim =2cm 0.5cm 2cm 1cm, clip = false, width=0.47\textwidth,height=0.27\textwidth]{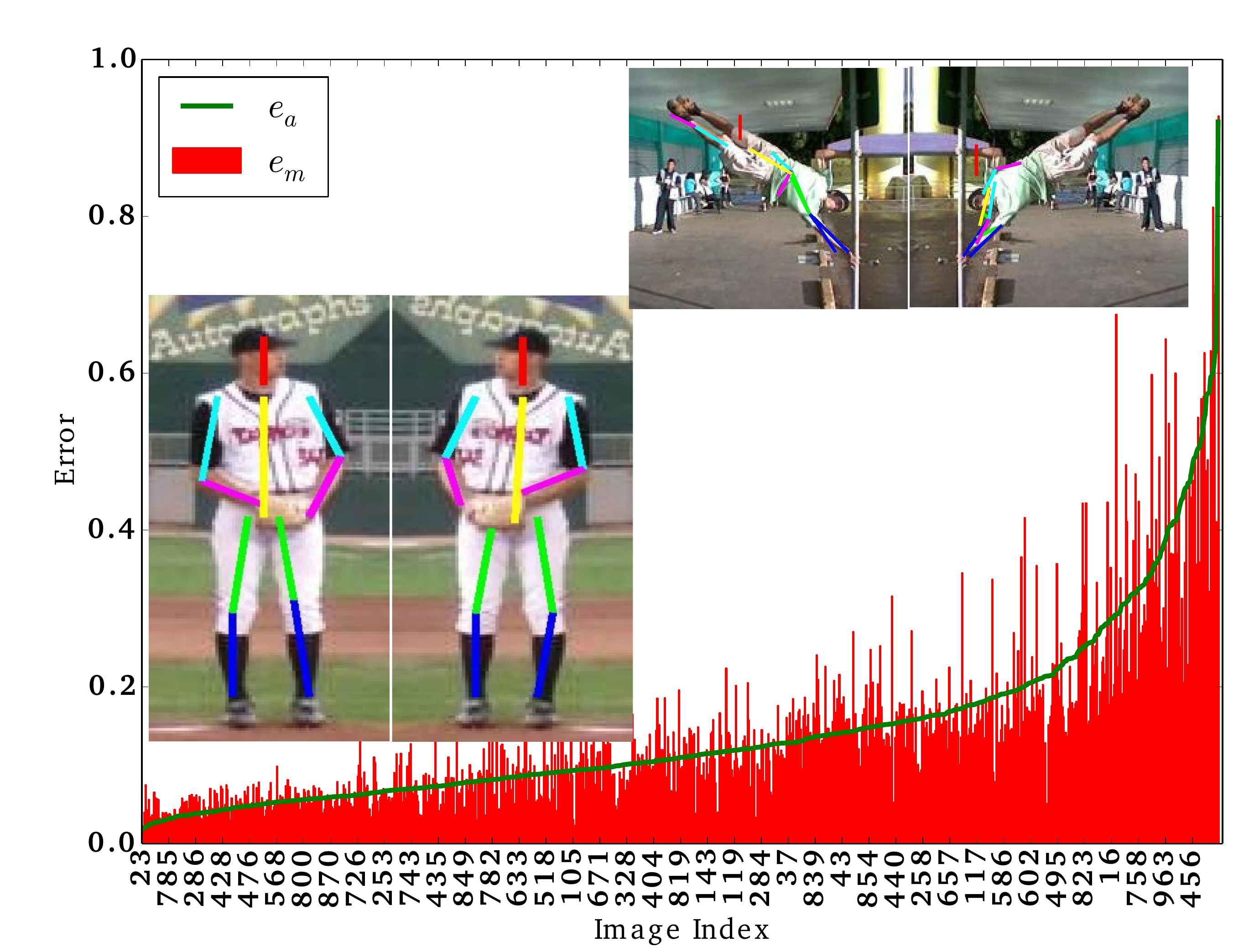}\\
\caption{Mirror error and alignment error on LSP of LTM \cite{fwang2013pose}. The $x$ axis is the image indexes after sorting the alignment error in ascend. Two example images and their mirror images are shown, one with small mirror error and the other with large mirror error.}
\label{fig::lspsortedmrr}
\end{figure}

\begin{figure}
\centering
\subfloat[{\scriptsize Yang \& Ramanan \cite{yang2013articulated} on Buffy}]{\includegraphics[trim =1cm 0.5cm 1cm 1cm, clip = false, width=0.23\textwidth,height=0.16\textwidth]{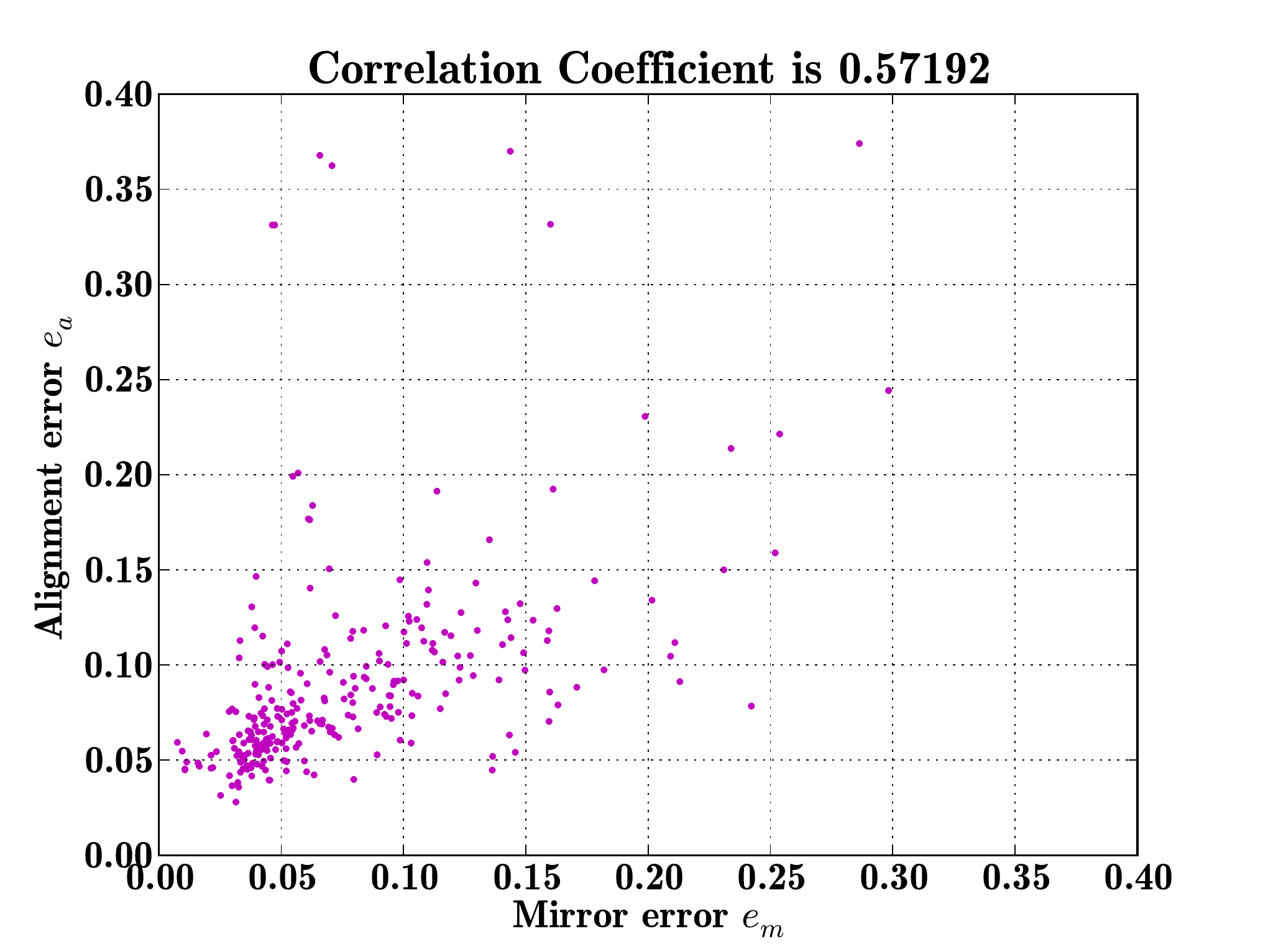}}
\subfloat[{\scriptsize Yang \& Ramanan \cite{yang2013articulated} on Parse}]{\includegraphics[trim =1cm 0.5cm 1cm 1cm, clip = false, width=0.23\textwidth,height=0.16\textwidth]{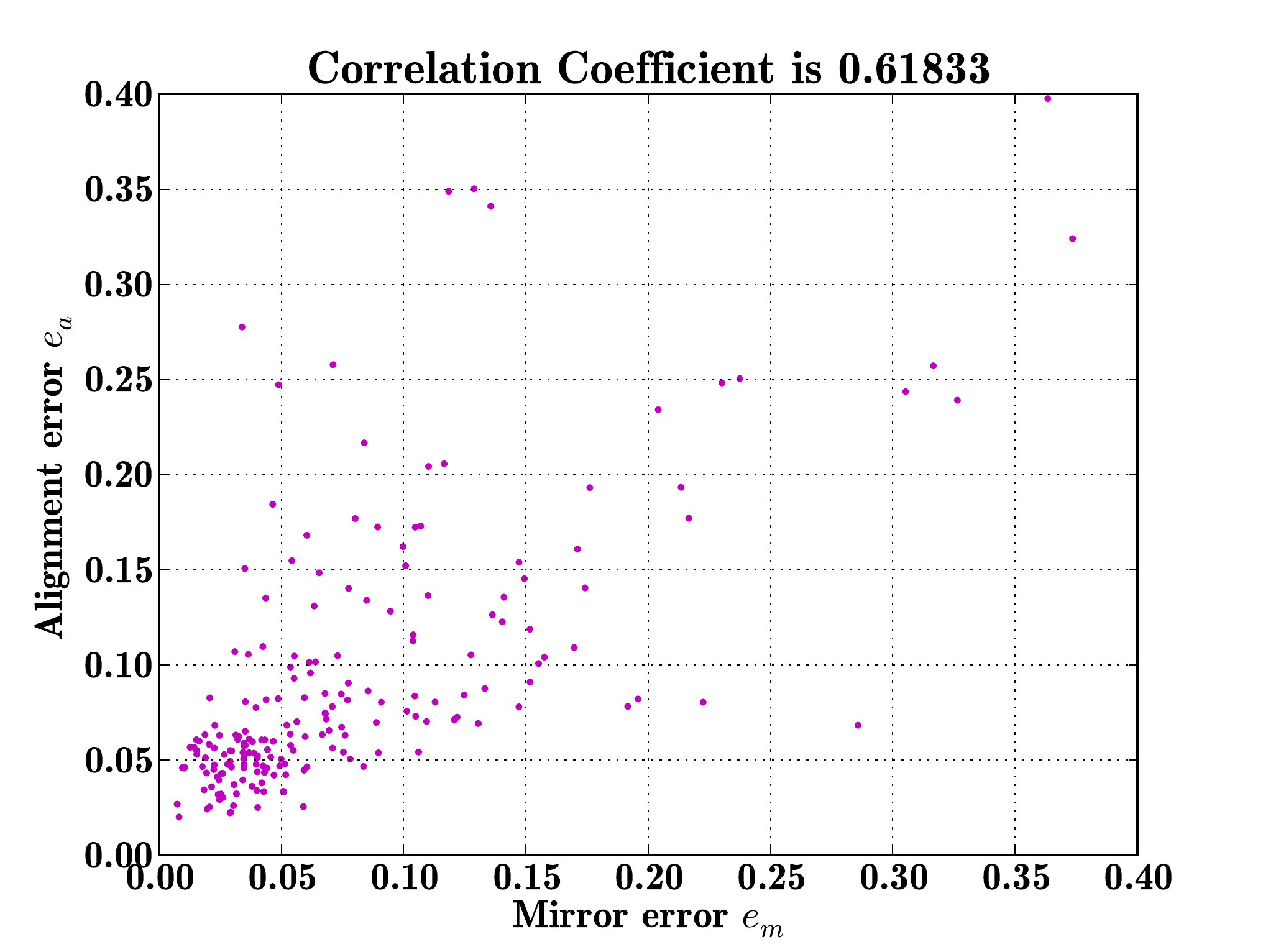}}\\
\subfloat[Yang and Ramanan \cite{yang2013articulated} on LSP]{\includegraphics[trim =1cm 0.5cm 1cm 1cm, clip = false, width=0.23\textwidth,height=0.16\textwidth]{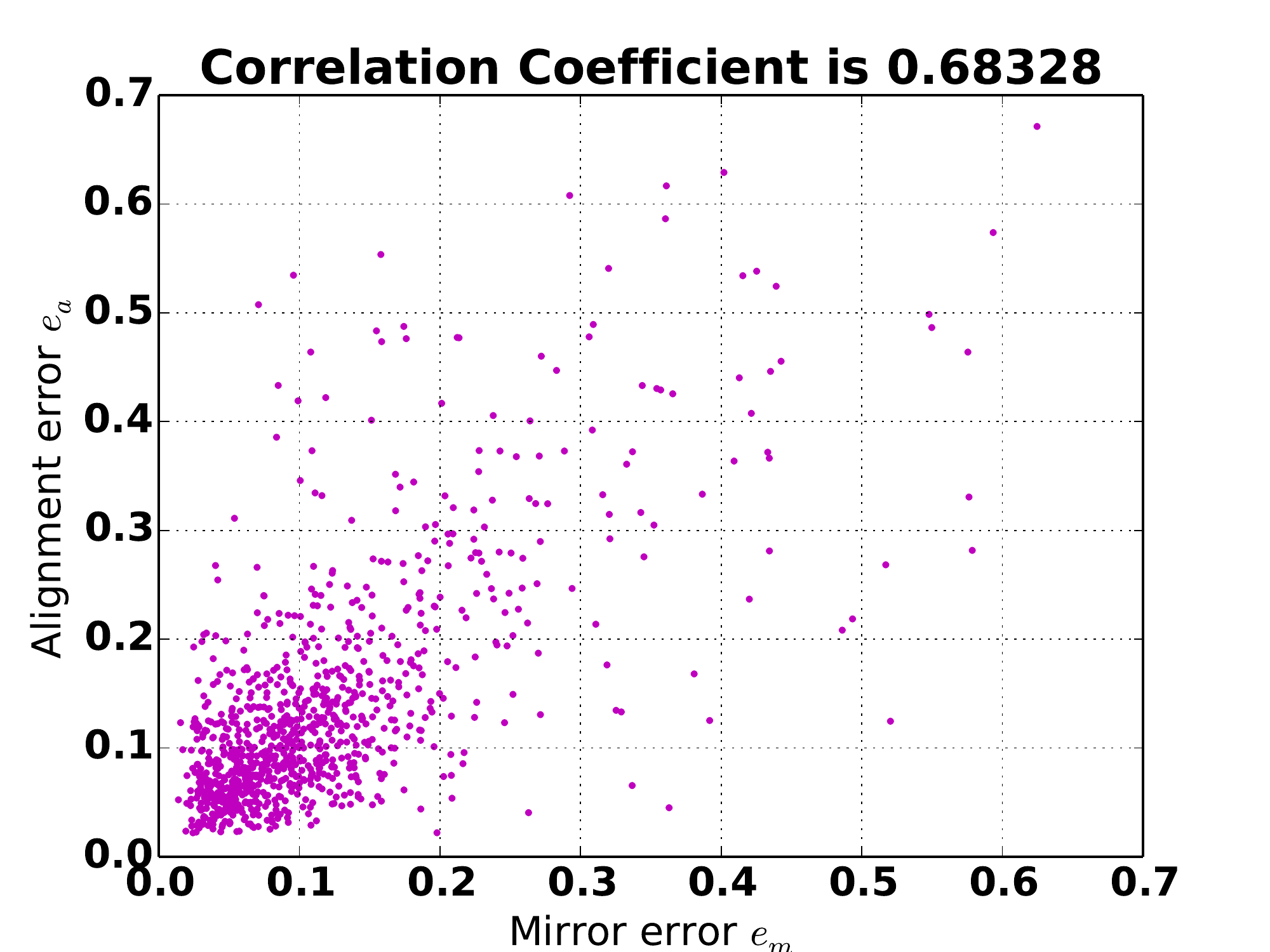}}
\subfloat[Wang and Li \cite{fwang2013pose} on LSP]{\includegraphics[trim =1cm 0.5cm 1cm 1cm, clip = false, width=0.23\textwidth,height=0.16\textwidth]{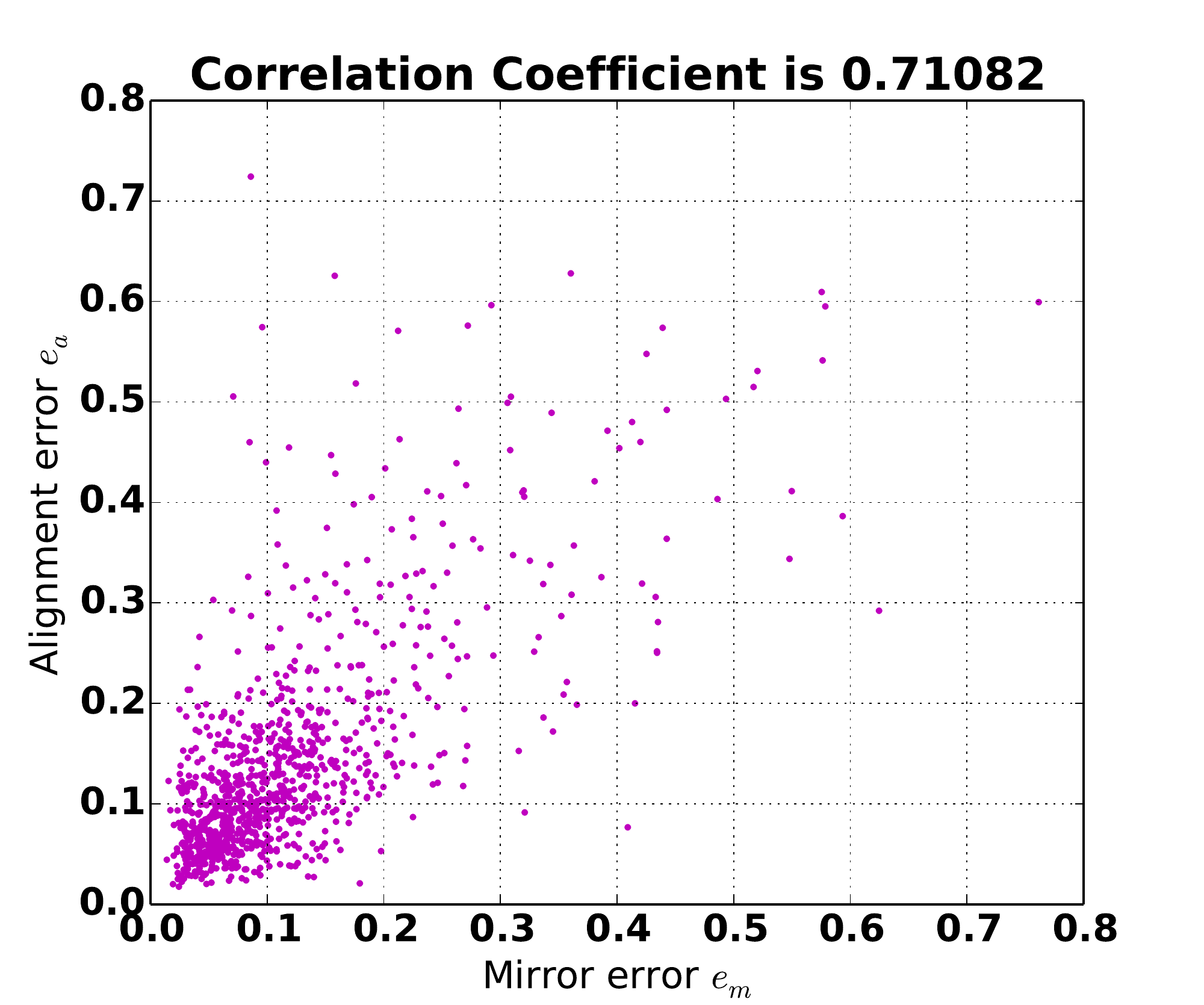}}
\caption{Correlation between the alignment error and mirror error. The correlation coefficients are shown above the figures.}
\label{fig::hprcorr}
\end{figure}

It is interesting to observe in Fig.~\ref{fig::visualizederror} that the joints with large average mirror error are usually the most challenging to localize, that is they are the ones with the higher alignment error. This seems to indicate that there is correlation between the mirror error and the alignment error. In Fig.~\ref{fig::lspsortedmrr}, as an example, we show the mirror error vs. the sorted sample-wise alignment error of LTM on LSP dataset. It is clear that the mirror error tends to increase as the image alignment error increases. Two examples of pairs of images are shown in Fig.~\ref{fig::lspsortedmrr} and the correlation between the sample-wise mirror error and the alignment error are shown them in Fig.~\ref{fig::hprcorr}.
On all three datasets the mirror error has shown a strong correlation to the alignment error. For the smaller datasets, Buffy and Parse the correlation coefficient is around 0.6. On the larger LSP dataset, the correlation coefficient of both LTM and  FMP is around 0.7. We can conclude that although the mirror error is calculated without knowledge of the ground truth, it is informative of the real alignment error in each sample. 

%

\subsection{Face alignment}

Face alignment has been intensively studied and most of the recent methods have reported close-to-human performance on face images "in the wild". Here, we look into the mirrorability of face alignment methods and how their error is correlated to the mirror error.

\paragraph{Experiment setting} For our analysis we focus on the most challenging datasets collected in the wild, namely the 300W. It is created for Automatic Facial Landmark Detection in-the-Wild Challenge \cite{sagonas300}. To this end, several popular data sets including LFPW \cite{belhumeur2011localizing}, AFW \cite{devacvpr2012face} and HELEN \cite{interactiveECCV2012} were re-annotated with 68 points mark-up and a new data set, called iBug, was added. We perform our analysis on a test set that comprises of the test images from HELEN (330 images), LFPW (224 images) and the images in the iBug subset (135 images), that is 689 images in total. The images in the iBug subset are extremely challenging due to the large head pose variations, faces that are partially outside the image and heavy occlusions. The test images are flipped horizontally to get the mirror images. We evaluate the performance of several recent state of the art methods, namely the Supervised Descent Method (\textbf{SDM}) \cite{xiong2013supervised}, the Robust Cascaded Pose Regression (\textbf{RCPR}) \cite{burgos2013robust}, the Incremental Face Alignment (\textbf{IFA}) \cite{asthanaincremental} and the Gaussian-Newton Deformable Part Model (\textbf{GN-DPM}) \cite{tzimiropoulos2014gauss}. For SDM, IFA and GN-DPM, only the trained models and the code for testing is available - we use those to directly apply them on the test images. As stated in the corresponding papers, the IFA and GN-DPM were trained on the 300W dataset and the SDM model was trained using a much larger dataset. SDM, IFA and GN-DPM only detect the 49 inner facial points - our analysis on those methods is therefore based on those points only.
For RCPR, for which the code for training is available, we retrain the model on the training images of 300W for the full 68 facial points mark-up. All those methods build on the result of a face detector - since most of them are sensitive to initialization, we carefully choose the \textit{right} face detector for each one to get the best performance. More specifically, for the IFA and GN-DPM we use the 300W face bounding boxes and for SDM and RCPR we use the Viola-Jones bounding boxes, that is for each method we used the detector that it used during training. For the methods that use the Viola-Jones bounding boxes, we checked manually to verify that the detection is correct - for those face images on which the Viola-Jones face detector fails, we adjust the 300W bounding box to roughly approximate the Viola-Jones bounding box.
\begin{figure}
\includegraphics[trim =1cm 0.5cm 1cm 1cm, clip = false, width=0.45\textwidth,height=0.27\textwidth]{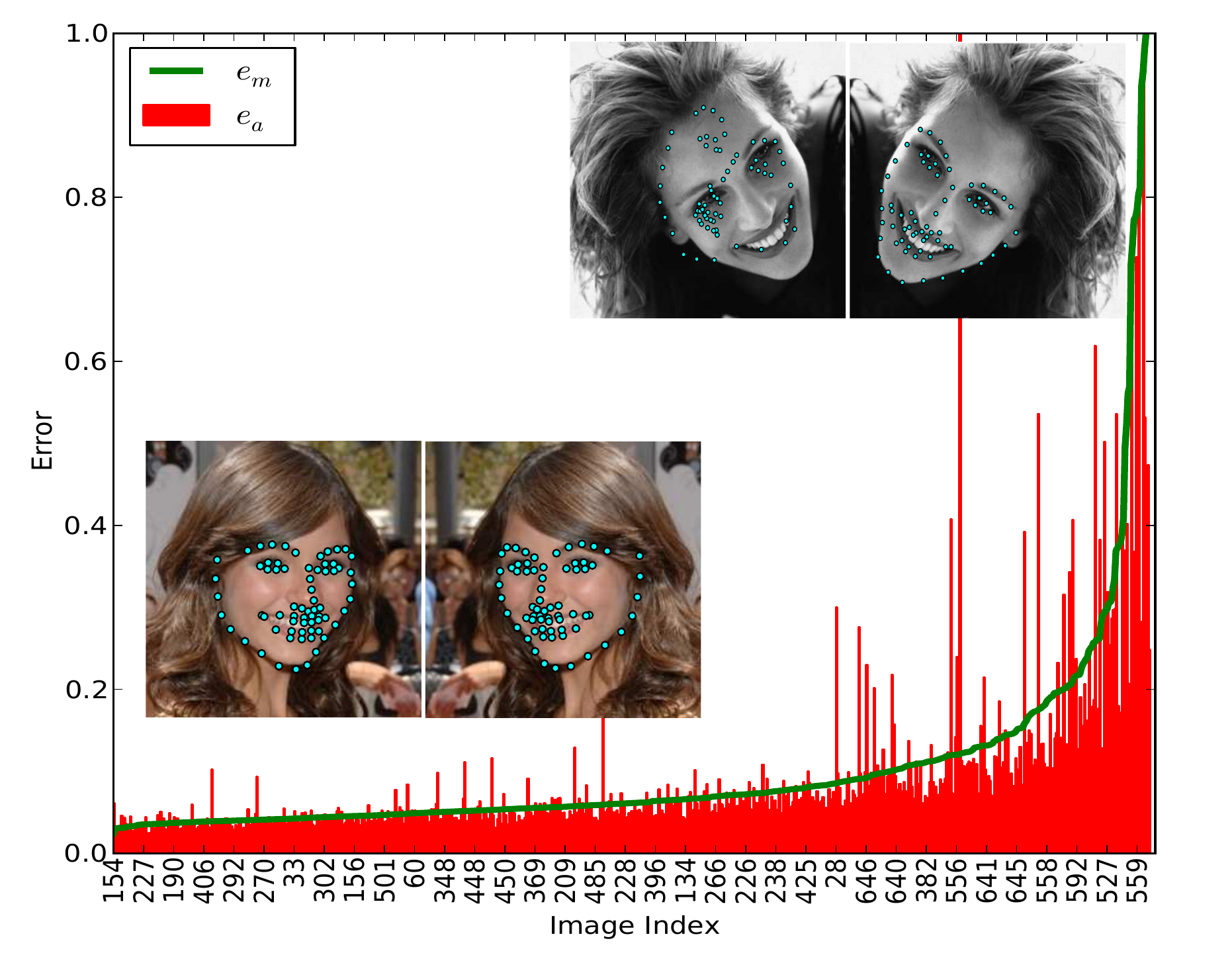}\\
\caption{Mirror error and alignment error of RCPR \cite{burgos2013robust} on 300W test images. Results are calculated over 68 facial points. }
\label{fig::rcprsortedmrr}
\end{figure}

\begin{figure}
\includegraphics[trim =1cm 0.5cm 1cm 1cm, clip = false, width=0.45\textwidth,height=0.27\textwidth]{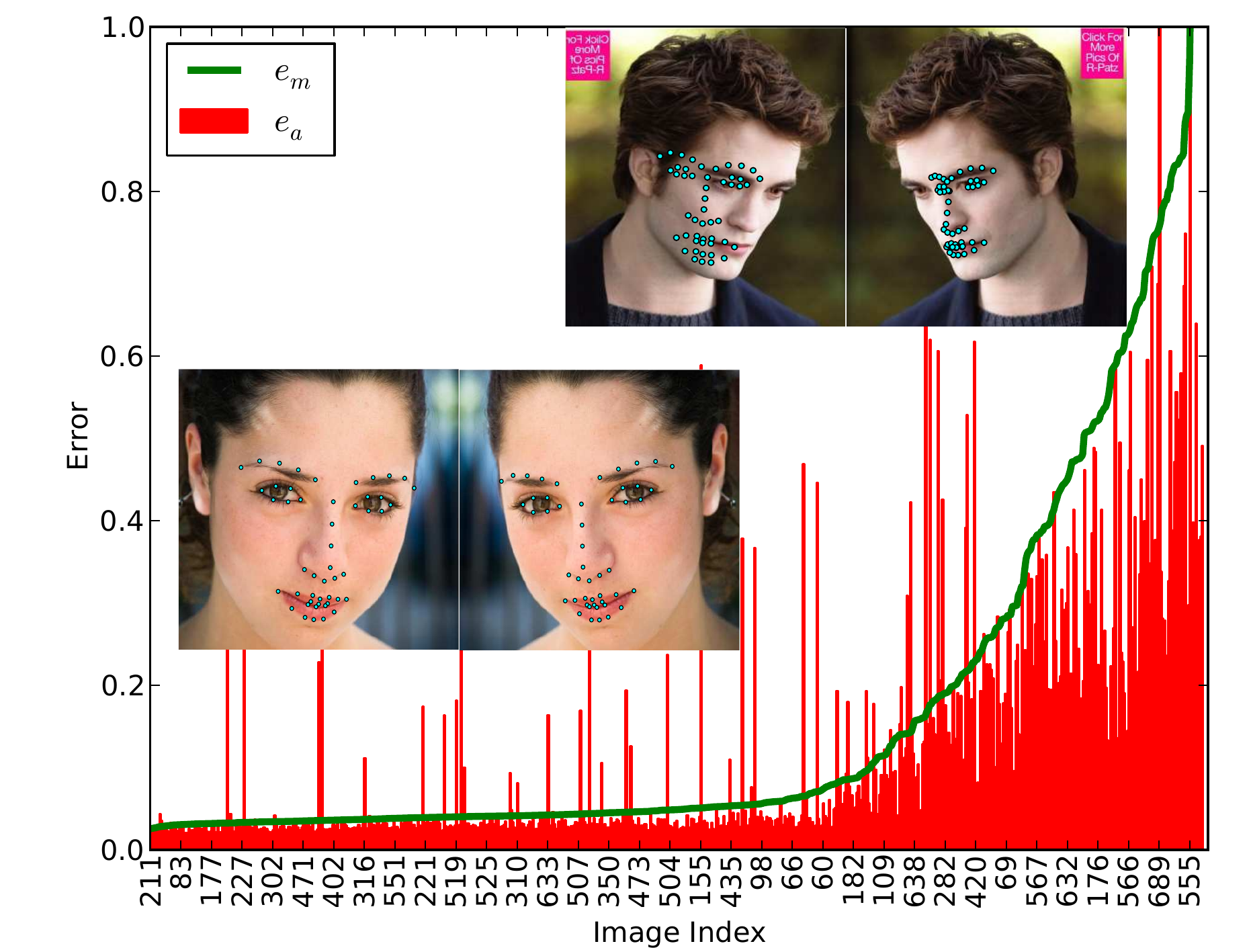}\\
\caption{Mirror error and alignment error of GN-DPM \cite{tzimiropoulos2014gauss} on 300W test images. Results are calculated over 49 inner facial points.}
\label{fig::gndpmsortedmrr}
\end{figure}
\begin{figure}
\centering
\subfloat[SDM \cite{xiong2013supervised}, 49P]{\includegraphics[trim =0cm 0cm 0cm 0cm, clip = true, width=0.23\textwidth,height=0.15\textwidth]{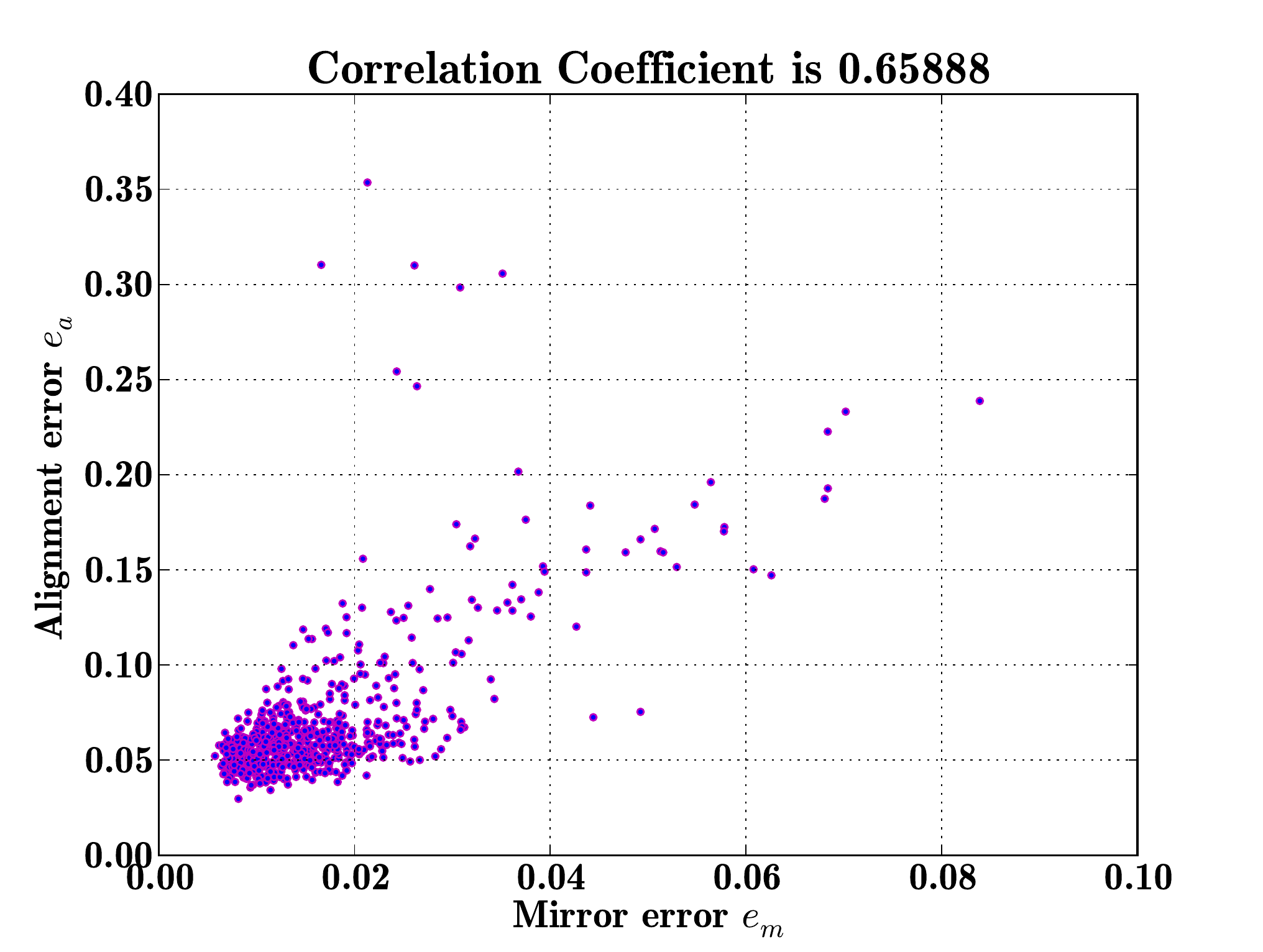}}
\subfloat[RCPR \cite{burgos2013robust}, 68P]{\includegraphics[trim =0cm 0cm 0cm 0cm, clip = true, width=0.23\textwidth,height=0.15\textwidth]{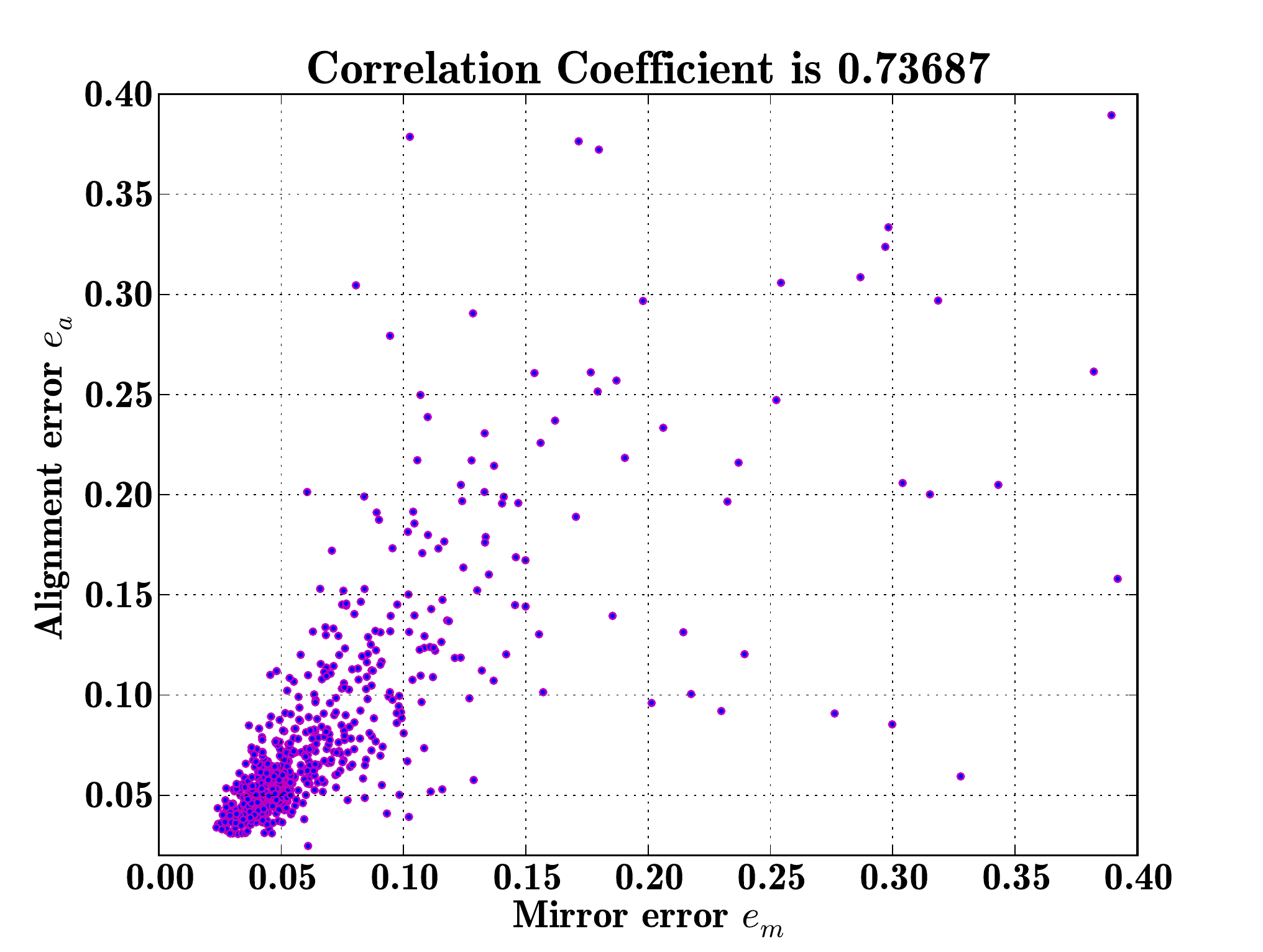}}\\
\subfloat[IFA \cite{asthanaincremental}, 49P]{\includegraphics[trim =0cm 0cm 0cm 0cm, clip = true, width=0.23\textwidth,height=0.15\textwidth]{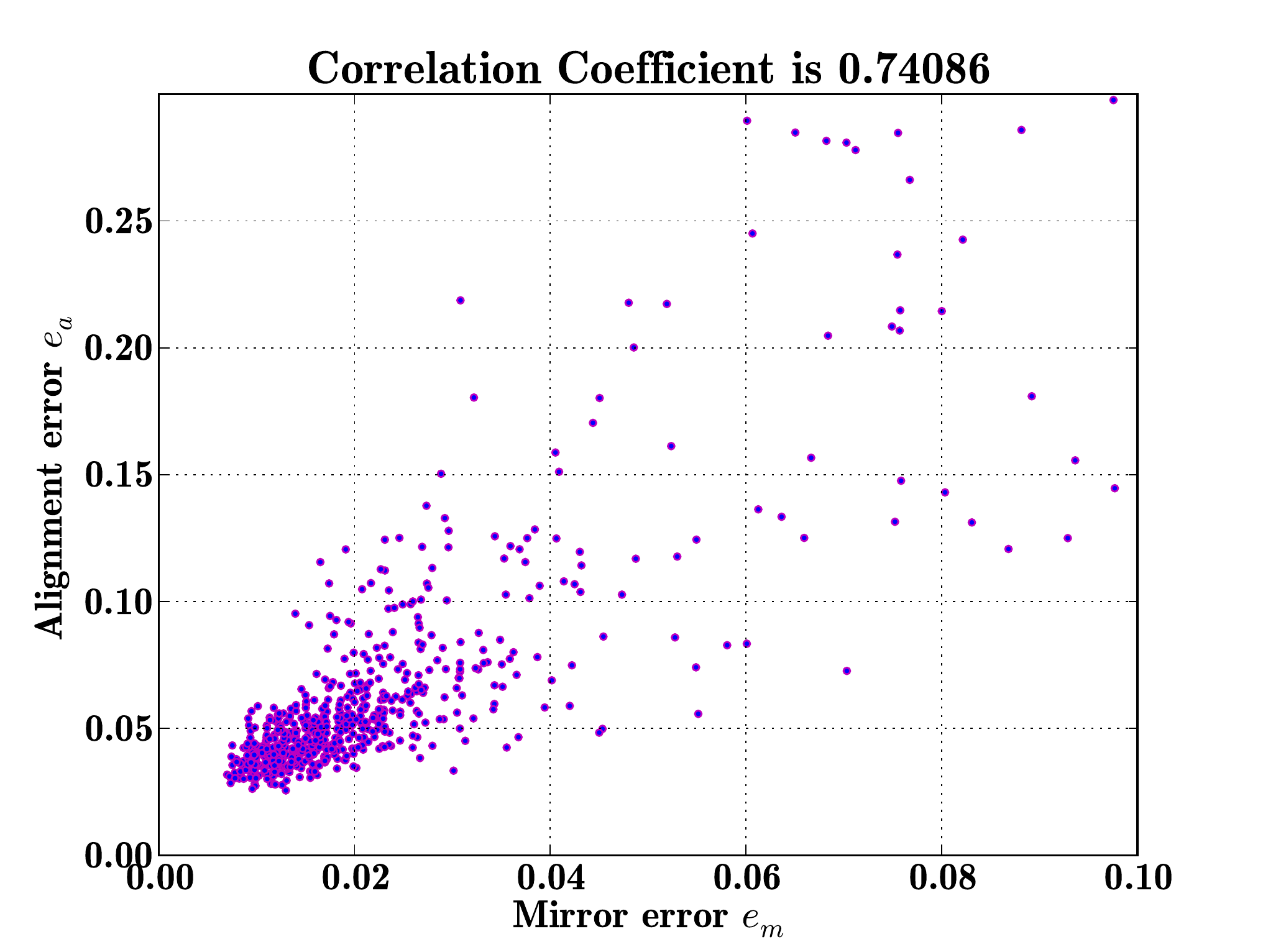}}
\subfloat[GN-DPM \cite{tzimiropoulos2014gauss}, 49P]{\includegraphics[trim =0cm 0cm 0cm 0cm, clip = true, width=0.23\textwidth,height=0.15\textwidth]{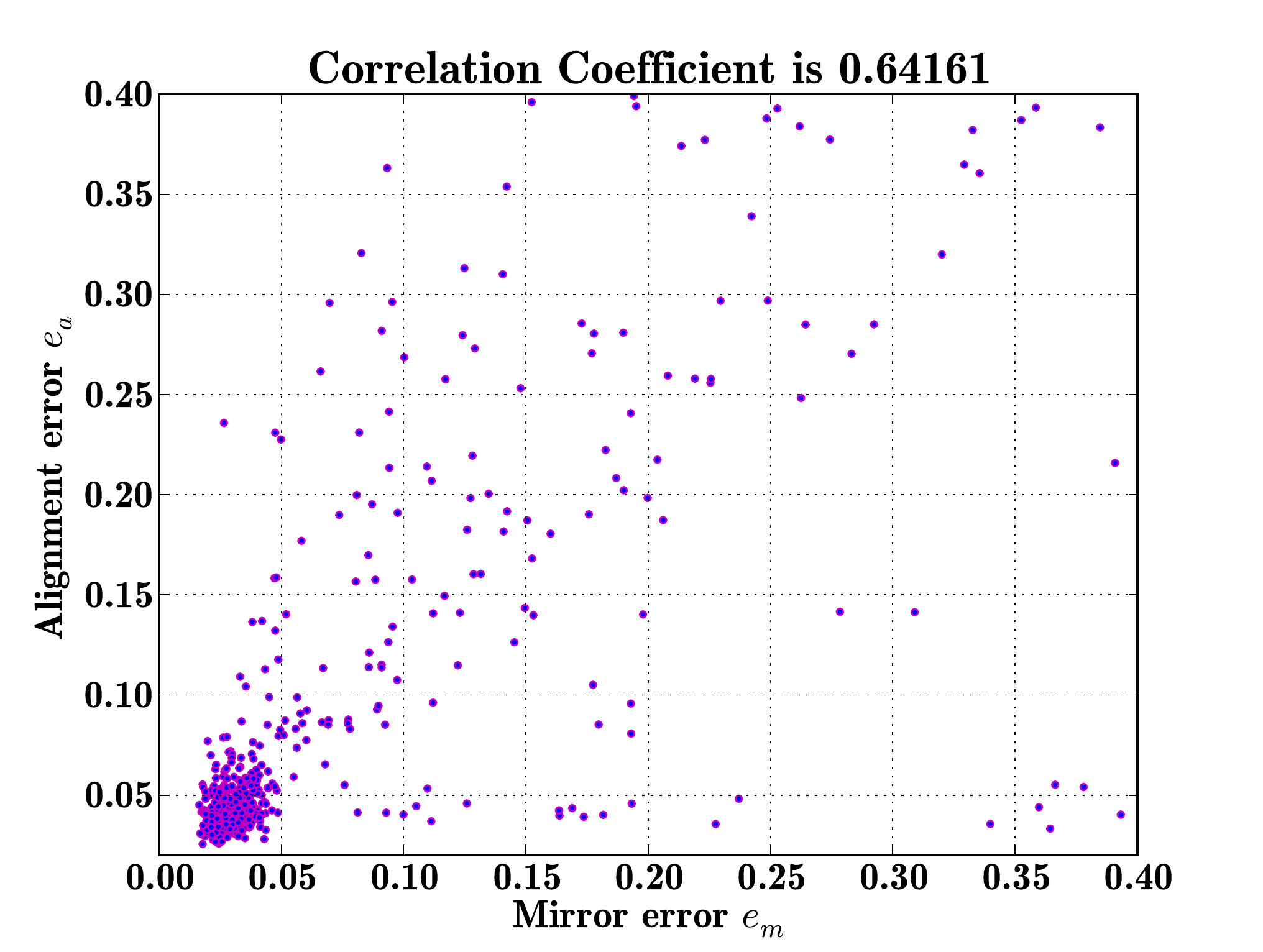}}
\caption{Correlation between the alignment error and the mirror error of various state of the art face alignment methods. The correlation coefficients are shown above the figures.}
\label{fig::facealignmentcorr}
\end{figure}

\paragraph{Mirrorability}
We calculated the mirror error and the alignment error for each of the 689 test samples in 300W for SDM, IFA, GN-DPM and RCPR. In Fig.~\ref{fig::gndpmsortedmrr} and Fig.~\ref{fig::rcprsortedmrr} we show the errors for two of the algorithms, i.e., the GN-DPM and the RCPR. The former is a representative local-based method and the latter a representative holistic-based method. Similar results were obtained for SDM and IFA. In each figure, two pairs of example images are shown - one with low mirror error (lower left corner) and one with large mirror error (upper right corner).  We sort the sample-wise alignment error in ascending order and plot it together with the corresponding sample mirror error. It is clear that although GN-DPM and the RCPR work in a very different way, for both the mirror error tends to increase as the alignment error increases. There are a few impulses in the lower range of the red curve, i.e., low $^\mathrm{q}e_a$ and high $e_m$. This means that although the algorithm has small alignment error on the original samples it has large error on the mirror images, i.e., $^\mathrm{q}e_a$ is high. There are three cases that result in high mirror error: 1) low $^\mathrm{q}e_a$ and high $^\mathrm{p}e_a$; 2) high $^\mathrm{q}e_a$ and low $^\mathrm{p}e_a$ (shown in Fig.~\ref{fig::rcprsortedmrr} upper right corner); 3) high $^\mathrm{q}e_a$ and high $^pe_a$ (shown in Fig.~\ref{fig::gndpmsortedmrr} upper right corner). Finally, in order to quantify this insight, we present the correlation between the mirror error and the alignment error in Fig.~\ref{fig::facealignmentcorr}. In all of the four methods there is a strong correlation between the mirror error and the alignment error with correlation coefficients ranging from 0.64 to 0.74 - these are very high.


\section{Mirrorability Applications}

In the previous sections we have shown that one of the nice properties of the mirror error is that it is strongly correlated with the object alignment error, that is with the ground truth error. In this section we show how it can be used in two practical applications, namely for selecting difficult samples and for providing feedback in a cascaded face alignment method.

\subsection{Difficult samples selection}


For any computer vision task, including face alignment, it is generally accepted that some samples are relatively more difficult than others, that is the error of the algorithm on them is higher. However, it is very difficult to estimate a measure of how well the algorithm has performed on a given sample without knowledge of the ground truth. Such a measure would be very useful, for example in order to select a proper alignment model for a given dataset or to select which samples to annotate in an Active Learning scheme. Here, we show how the mirror error can be used for selecting difficult samples in the problem of face alignment. In order to do so we apply several methods (IFA, SDM, GN-DPM, RCPR) on the test images of the 300W and get the detection results. Then we sort the normalized mirror error $e_m$ in descending order and select the first $M$ samples as being the most difficult ones. We denote this set as $S_{e_m}$. 
\begin{figure*}
\centering
\subfloat[$\rho$ of $S_{e_a} \Leftrightarrow S_{e_m}$. ]{\includegraphics[trim =0cm 0cm 0cm 0cm, clip = true, width=0.33\textwidth,height=0.23\textwidth]{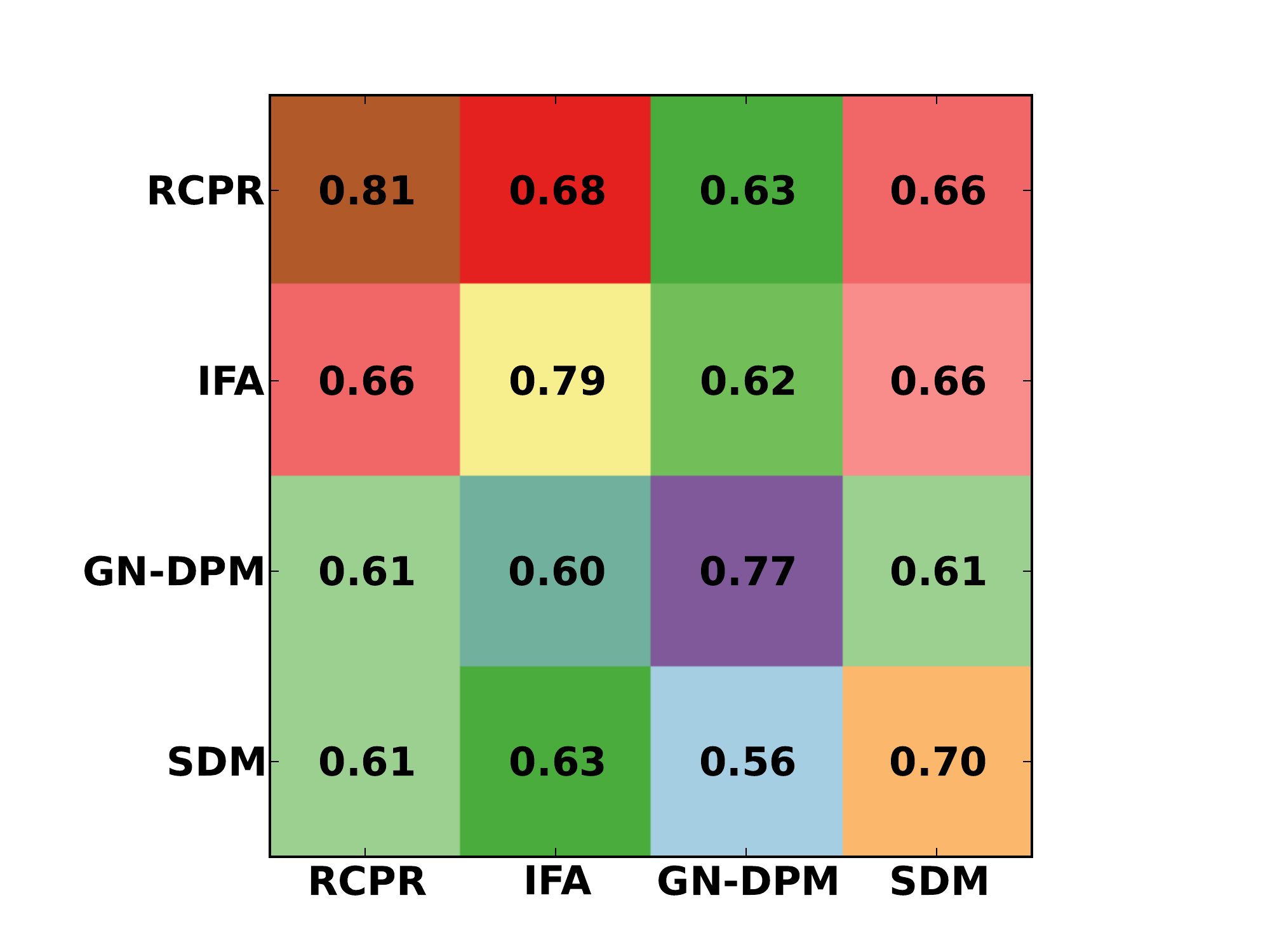} \label{fig::confusionm1} }
\subfloat[$\rho$ of $S_{e_m} \Leftrightarrow S_{e_m}$.]{\includegraphics[trim =0cm 0cm 0cm 0cm, clip = true, width=0.33\textwidth,height=0.23\textwidth]{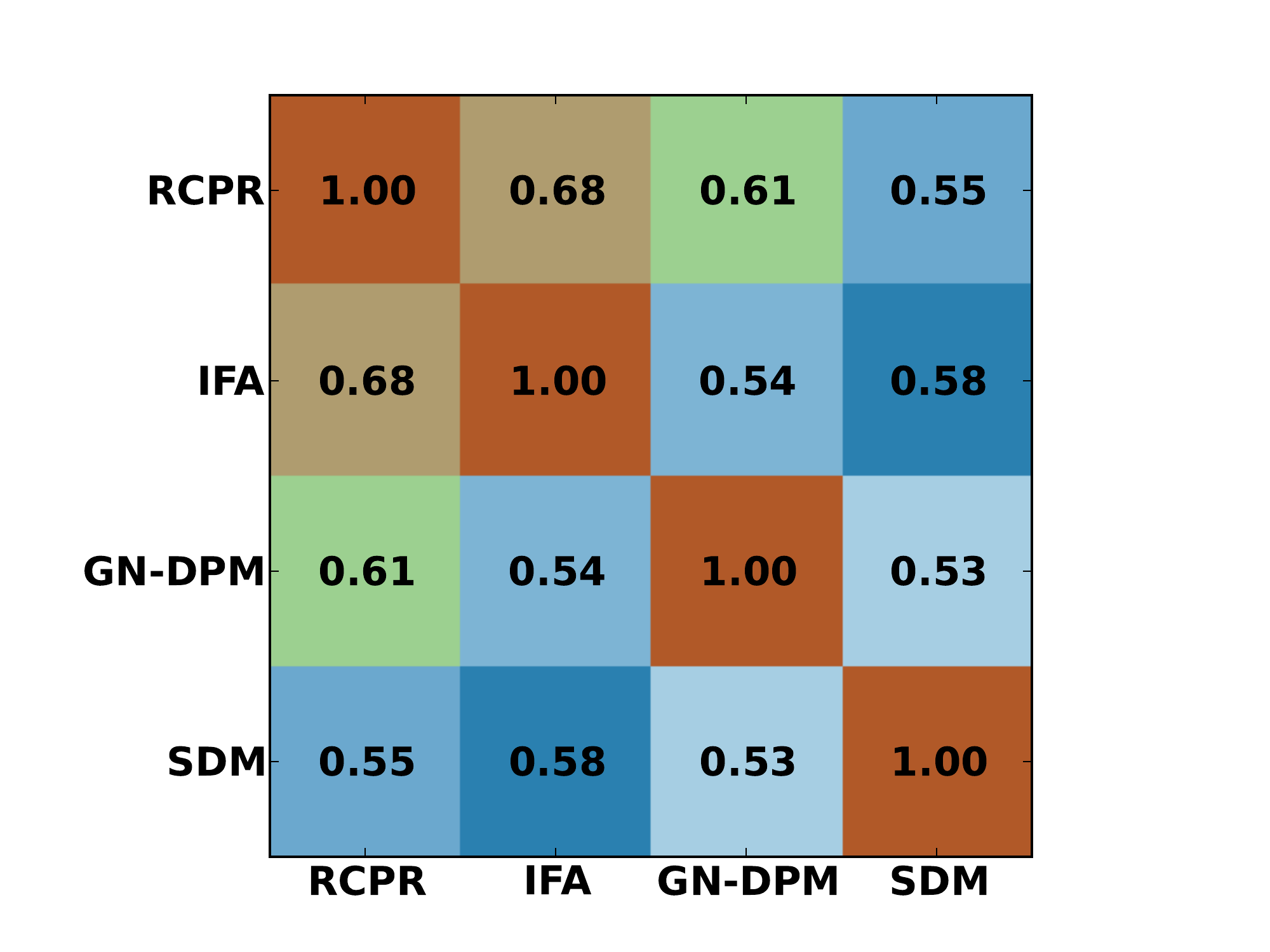}\label{fig::confusionm2}}
\subfloat[$\rho$ of $S_{e_a} \Leftrightarrow S_{e_a}$.]{\includegraphics[trim =0cm 0cm 0cm 0cm, clip = true, width=0.33\textwidth,height=0.23\textwidth]{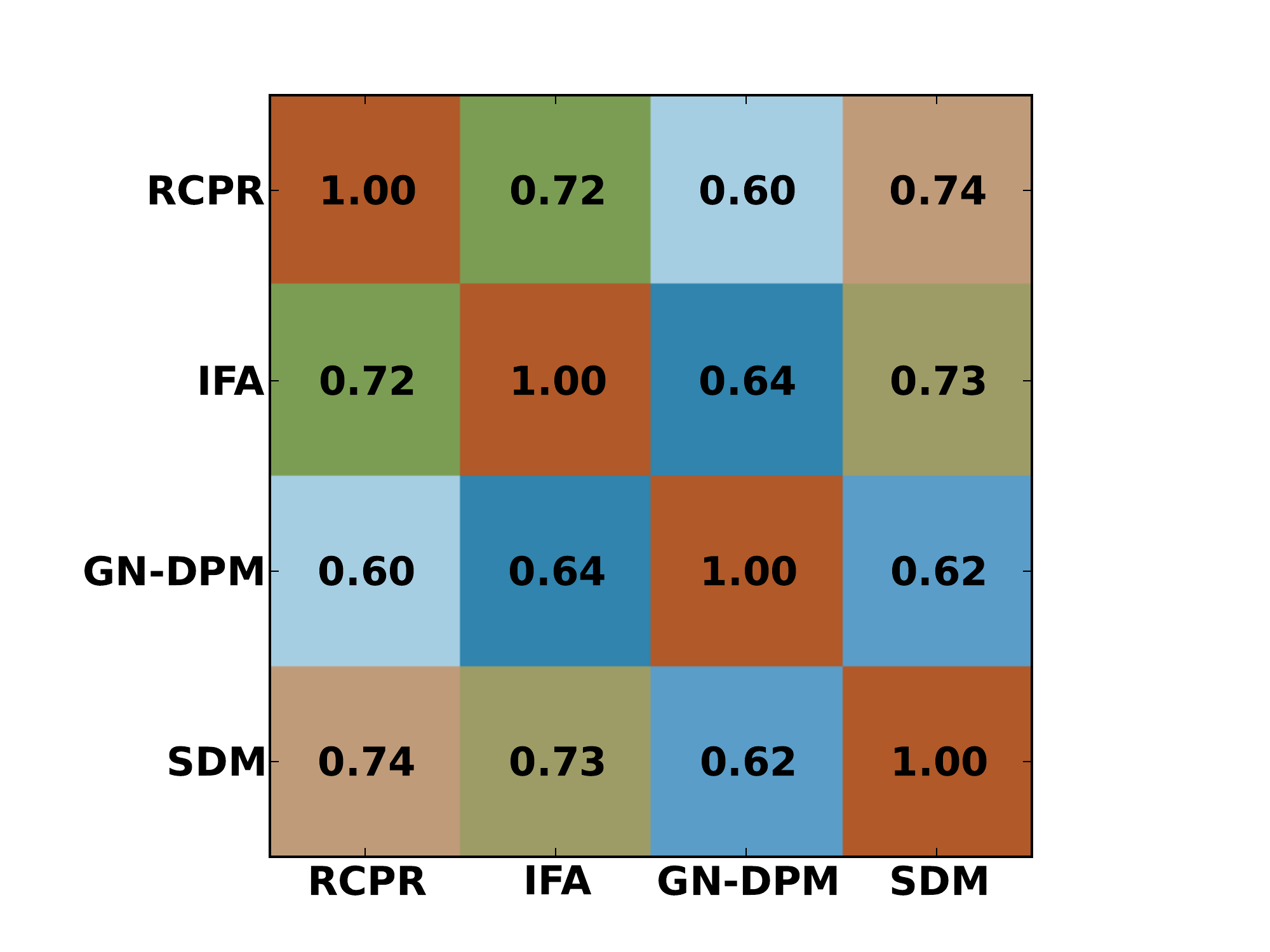}\label{fig::confusionm3}}
\caption{Consistency measure of 'difficult' samples detection, with $M=150$.   }
\label{fig::confusionmatrix}
\end{figure*}

In order to evaluate whether the samples that we have selected in this way are truly 'difficult' we measure  the similarity between the set containing those $M$ selected samples and the set $S_{e_a}$ that contains the $M$ samples that have the largest alignment error $e_a$ for each method. We use a measure that we call consistency which we define as the fraction of the common samples between the two sets, that is
\begin{equation}
\rho = \frac{|S_1\cap S_2|}{M}
\label{eq::pairwiseconsistency}
\end{equation}
where $|S_1\cap S_2|$ is the size of the intersection of $S_1$ and $S_2$. For each method $i$, we calculate two sets each containing $M$ samples, i.e., $S^i_{e_m}$ and $S^i_{e_a}$. We set the value of $M$ to 150.  The chance rate is $\frac{M}{N}$, where $M$ is the number of selected and $N$ is the size of the dataset -  in our case is $\frac{150}{689} \approx 0.22$.

 The pairwise consistency rate matrix of $S^i_{e_m}$ and $S^i_{e_a}$ is shown in Fig.~\ref{fig::confusionm1}, where in a certain row we show the consistency between the $S^i_{e_m}$ of a certain method with the $S^i_{e_a}$ of all methods, including the method itself. Note that the diagonal does not contain ones, since $S^i_{e_m}$ are the $M$ samples with the highest mirror error and $S^i_{e_a}$ the $M$ samples with the highest alignment error. As it can be seen, the consistency between the two sets of samples for a specific method (i.e., the diagonal values) are all above 0.7 - the highest is 0.81 for RCPR. More interestingly, the consistency across different methods, i.e., the $M$ samples selected according to $e_a$ for a method in a certain row and the $M$ samples selected according to $e_m$ in a certain column is high, with values ranging from 0.56 to 0.68.  This shows that the samples that we have selected are truly 'difficult', not only for the method employed in the selection process but also for the other face alignment methods. In other words this shows that the methods that we have examined have difficulties with the same images.

Second, we evaluate the consistency across different approaches, i.e., the consistency of 'difficult' samples found by different approaches. Thus, we calculate the pairwise consistency of $S^i_{e_m}$ of those methods as shown in Fig.~\ref{fig::confusionm2}.  The resulting values are clearly much higher than the chance value of 0.22. In Fig.~\ref{fig::confusionm3} we depict the 'optimal' case where the ground truth, that is the alignment error itself, is used to calculate the pairwise consistency. We observe that the consistency calculated by our selection process is very close to the one calculated based on the ground truth.  We can further conclude that:
\begin{itemize}
\item the difficulty of samples is shared by the different methods that we have examined.
\item the difficult samples selected by the mirror error show high consistency across different approaches. 
\end{itemize}

\subsection{Feedback on cascaded face alignment}
In recent years cascaded methods like SDM \cite{xiong2013supervised}, IFA \cite{asthanaincremental}, CFAN \cite{cfaneccv2014} and RCPR \cite{burgos2013robust} have shown promising results in face alignment. Although they differ in terms of the regressor and the features that they use in each iteration they all follow the same strategy. The methods start from one or several initializations of the face shape, that are often calculated from the face bounding box, and then iteratively refine the estimation of the face shape by applying at each iteration a regressor that estimates the udpate of the shape. These methods are intrinsically sensitive to the initialization \cite{burgos2013robust,cfaneccv2014} . As stated in \cite{xiongSDMarxiv}, only initializations that are in a range of the optimal shape can converge to the correct solution. To address this problem, \cite{suncvpr2012} proposed to use several random initializations and give the final estimate as the median of the solutions to which they convergence. However, having several randomly generated initializations does not guarantee that the correct solution is reached. The 'smart restart' proposed in \cite{burgos2013robust} has improved the results to a certain degree. The scheme starts from different initializations and apply only 10\% of the cascade. Then, the variance between the predictions is checked. If the variance is below a certain threshold, the remaining 90\% of the cascade is applied as usual. Otherwise the process is restarted with a different set of initializations. 

Here, we propose to use the mirror error as a feedback to close this \textit{open} cascaded system. More specifically, for a given test image we first create its mirror image. Then we apply the RCPR model on the original test image and the mirror image and calculate the mirror error. If the mirror error is above a threshold we restart the process using different initializations, otherwise we keep the detection results. This procedure can be applied until the mirror error is below a threshold, or until a  maximum number of iterations $M$ is reached. 
In contrast to the original RCPR method that keeps only the results from the last set of initializations, we keep the one that has the smallest mirror error. This makes sense since new random initializations do not necessarily lead to better results than past initializations.  

\begin{figure}
\centering
\subfloat[Original RCPR restart scheme. Presion=0.25, Recall = 0.63.]{\includegraphics[trim =9cm 0cm 6cm 0cm, clip = true, width=0.5\textwidth,height=0.12\textwidth]{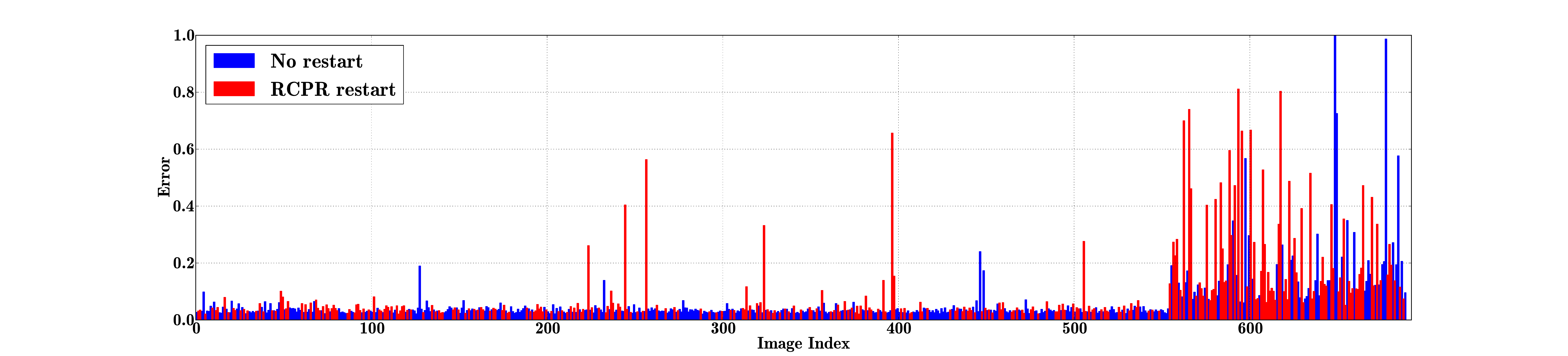} \label{fig::rcpr_restart_samples} }\\
\subfloat[Our restart scheme. Precision = 0.65, Recall = 0.63.]{\includegraphics[trim =9cm 0cm 6cm 1cm, clip = true, width=0.5\textwidth,height=0.12\textwidth]{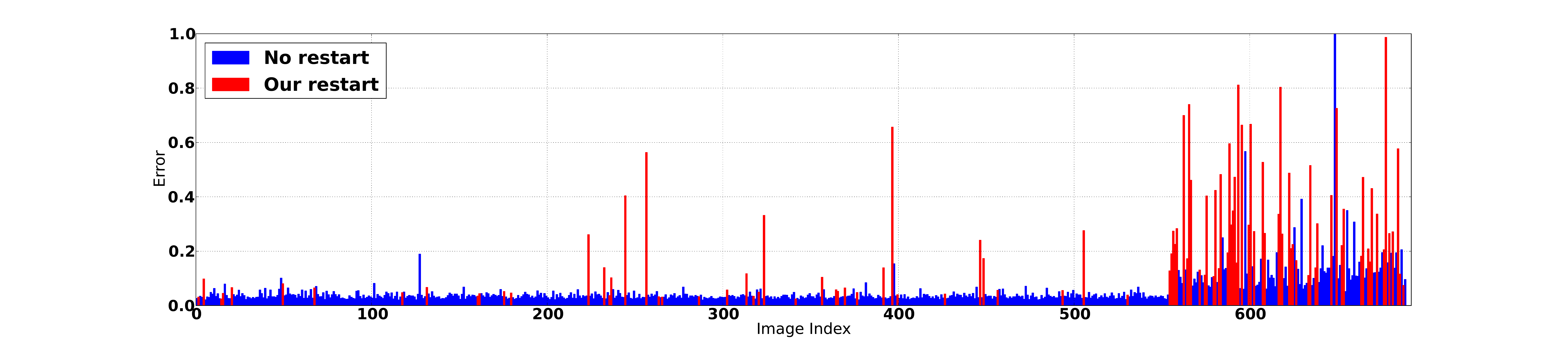}\label{fig::proposed_restart}}
\caption{Restart scheme of our method vs. RCPR \cite{burgos2013robust} (best viewed in color). }
\label{fig::restart}
\end{figure}

First we evaluate the effectiveness of  our feedback scheme. Ideally, the restart will be initiated only when the current initialization is unable to lead to a \textit{good} solution. Treating it as a two class classification problem we report results using a precision-recall based evaluation. A face alignment is considered to belong to the 'good' class if the mean alignment error is below $10\%$ of the inter-ocular distance, otherwise, it is considered to belong to the 'bad' class - in the latter case a re-start is needed. The precision is the number of samples classified correctly as belonging to the 'bad' (positive) class divided by the total number of samples that are classified as belonging to the 'bad' class.  Recall in this context is defined as the number of true positives divided by the total number of samples that belong to the bad class. For a fair comparison, we adjust our threshold on the mirror error (i.e. the threshold above which we restart the cascade with a different initialization) to get similar recall as the RCPR with smart re-start \cite{burgos2013robust} gets using its default parameters. We note that our parameter can also be optimized by cross validation for better performance. As can be seen in Fig.~\ref{fig::restart}, at a similar recall level, our proposed scheme has significantly higher precision (0.65 vs. 0.25) than that of RCPR 'smart re-start', this verifies that our method is more effective in selecting samples for which restarting initializations are needed.

Second, we evaluate the improvement in the face alignment that we obtain using our proposed feedback scheme. We compare to 1) RCPR without restart (RCPR-O), 2) RCPR with the smart restart of \cite{burgos2013robust} (RCPR-S) and 3) other state of the art methods. We create two versions of our method. The first version, RCPR-F1, uses 5 initializations and at most two restarts - this allows direct comparison to the baseline method that uses the same number of initializations and restarts. The second version, RCPR-F2, uses 10 initializations and at most 4 times of restarts - this version produces better results and still has good runtime performance. We compare to SDM \cite{xiong2013supervised}, IFA \cite{asthanaincremental}, GN-DPM \cite{tzimiropoulos2014gauss} and CFAN \cite{cfaneccv2014} - all of those have publicly available software and report good results. The results of the comparison is shown in Table~\ref{tab::facealignmentperformance}.  We compare the normalized alignment error of the common 49 inner facial landmarks for all of these methods and the 68 facial landmarks whenever this is possible. On the challenging 300W test set, with our proposed feedback scheme, the RCPR method has the best performance compared to not only the original version of RCPR but also to all the other methods. Although good performance is obtained on the face alignment problem, we emphasize that the main focus of this work is to bring attention to the mirroability of object localization models.
\begin{table}
\centering
\scriptsize
\setlength{\tabcolsep}{2.5pt}
\begin{tabular}{c|cc|cc|cccc}
\hline
Methods&\textit{RCPR-F2}&\textit{RCPR-F1}&RCPR-S&RCPR-O&SDM&IFA&GN-DPM&CFAN\\
49P&\textbf{5.35}&6.07&6.59&7.14&7.12&8.31&12.42&7.24\\
68P&\textbf{6.25}&7.11&7.42&7.73&-&-&-&7.72\\
\hline
\end{tabular}
\caption{49/68 facial landmark mean error comparison .}
\label{tab::facealignmentperformance}
\end{table}

\section{Related Work}

As a method that estimates the quality of the output of a vision system, our method is related to works like the meta-recognition \cite{scheirer2011meta}, face recognition score analysis \cite{wang2007modeling} and the recent failure alert \cite{zhangpredicting} for failure prediction. Our method differs from those works in two prominent aspects (1) we focus on fine-grained object part localization problem while they focus on instance level recognition or detection. (2) we do not train any additional models for evaluation while all those methods rely on meta-systems. In the specific application of evaluating the performance of Human Pose Estimation, \cite{jammalamadaka2012has} proposed an evaluation algorithm, however, again such an evaluation requires a meta model and it only works for that specific application. 

Our method is also very different from object/feature detection methods that exploit mirror symmetry as a constraint in model building \cite{tsogkas2012learning,loy2006detecting}. We note that our model does not assume that the detected object or shape appears symmetrically in an image - such an assumption clearly does not hold true for the articulated (human body) and deformable (face) objects that we are dealing with. None of the methods that we have exploited in this paper explicitly used the appearance symmetry in model learning.  Our method only utilizes the mirror symmetry property to map the object parts between the original and mirror images. 
 
Developing transformation invariant vision system has drawn much attention in the last decades. Examples are the rotation invariant face detection method \cite{rowley1998rotation} and the scale invariant feature transform (SIFT) \cite{lowe1999object}, which handle efficiently several transformations including the mirror transformation. Recently, Gens and Domingos proposed the Deep Symmetry Networks \cite{gensdeep} that use symmetry groups to represent variations - it is unclear though how the proposed method can be applied for object part localization. Szegedy \textit{et al.} \cite{szegedy2013intriguing} has studied some intriguing properties of neural networks when dealing with certain artificial perturbations.  Our method focuses on examining the performance of object part localization methods on one of the simplest transforms, i.e. mirror transformation, and drawing useful conclusions.

\section{Conclusion and Discussion}

In this work, we have investigated how state of the art object localization methods behave on mirror images in comparison to how they behave on the original ones. Surprisingly, all of the methods that we have evaluated on two representative problems, struggle to get mirror symmetric results despite the fact that they were trained with datasets that were augmented with the mirror images.

In order to qualitatively analyze their behavior, we introduced the concept of mirrorability and defined a measure called the mirror error. Our analysis let to some interesting findings in mirrorability, among which a high correlation between the mirror error and ground truth error. Further, since the ground truth is not needed to calculate the mirror error, we show two applications, namely difficult samples selection and cascaded face alignment feedback that aids a re-initialization scheme. We believe there are many other potential applications in particular in Active Learning. 

The findings of this paper raise several interesting questions. Why some methods have shown better performance in terms of absolute mirror error, for example SDM is smaller and RCPR is bigger?  Can the design of algorithms with low mirrorability error lead to algorithms with good overall performance? We believe these are all interesting research problems for future work. 

{\small
\bibliographystyle{ieee}
\footnotesize
\bibliography{hyconference}

\begin{thebibliography}{10}\itemsep=-1pt

\bibitem{andriluka14cvpr}
M.~Andriluka, L.~Pishchulin, P.~Gehler, and B.~Schiele.
\newblock 2d human pose estimation: New benchmark and state of the art
  analysis.
\newblock In {\em CVPR}, 2014.

\bibitem{asthanaincremental}
A.~Asthana, S.~Zafeiriou, S.~Cheng, and M.~Pantic.
\newblock Incremental face alignment in the wild.
\newblock In {\em CVPR}, 2014.

\bibitem{belhumeur2011localizing}
P.~N. Belhumeur, D.~W. Jacobs, D.~J. Kriegman, and N.~Kumar.
\newblock {Localizing parts of faces using a consensus of exemplars}.
\newblock In {\em CVPR}, 2011.

\bibitem{burgos2013robust}
X.~P. Burgos-Artizzu, P.~Perona, and P.~Doll{\'a}r.
\newblock Robust face landmark estimation under occlusion.
\newblock In {\em ICCV}, 2013.

\bibitem{suncvpr2012}
X.~Cao, Y.~Wei, F.~Wen, and J.~Sun.
\newblock {Face alignment by explicit shape regression}.
\newblock In {\em CVPR}, 2012.

\bibitem{eichner20122d}
M.~Eichner, M.~Marin-Jimenez, A.~Zisserman, and V.~Ferrari.
\newblock 2d articulated human pose estimation and retrieval in (almost)
  unconstrained still images.
\newblock {\em IJCV}, 99(2):190--214, 2012.

\bibitem{finnerty2005did}
J.~R. Finnerty.
\newblock Did internal transport, rather than directed locomotion, favor the
  evolution of bilateral symmetry in animals?
\newblock {\em BioEssays}, 27(11):1174--1180, 2005.

\bibitem{gensdeep}
R.~Gens and P.~Domingos.
\newblock Deep symmetry networks.
\newblock In {\em NIPS}, 2014.

\bibitem{jammalamadaka2012has}
N.~Jammalamadaka, A.~Zisserman, M.~Eichner, V.~Ferrari, and C.~Jawahar.
\newblock Has my algorithm succeeded? an evaluator for human pose estimators.
\newblock In {\em ECCV}. Springer, 2012.

\bibitem{interactiveECCV2012}
V.~Le, J.~Brandt, Z.~Lin, L.~Bourdev, and T.~S. Huang.
\newblock {Interactive facial feature localization}.
\newblock In {\em ECCV}, 2012.

\bibitem{lowe1999object}
D.~G. Lowe.
\newblock Object recognition from local scale-invariant features.
\newblock In {\em ICCV}, 1999.

\bibitem{loy2006detecting}
G.~Loy and J.-O. Eklundh.
\newblock Detecting symmetry and symmetric constellations of features.
\newblock In {\em ECCV}. 2006.

\bibitem{ramanan2006learning}
D.~Ramanan.
\newblock Learning to parse images of articulated bodies.
\newblock In {\em NIPS}, 2006.

\bibitem{rowley1998rotation}
H.~A. Rowley, S.~Baluja, and T.~Kanade.
\newblock Rotation invariant neural network-based face detection.
\newblock In {\em CVPR}, 1998.

\bibitem{sagonas300}
C.~Sagonas, G.~Tzimiropoulos, S.~Zafeiriou, and M.~Pantic.
\newblock 300 faces in-the-wild challenge: The first facial landmark
  localization challenge.
\newblock In {\em ICCV}, 2013.

\bibitem{scheirer2011meta}
W.~J. Scheirer, A.~Rocha, R.~J. Micheals, and T.~E. Boult.
\newblock Meta-recognition: The theory and practice of recognition score
  analysis.
\newblock {\em T-PAMI}, 33(8):1689--1695, 2011.

\bibitem{szegedy2013intriguing}
C.~Szegedy, W.~Zaremba, I.~Sutskever, J.~Bruna, D.~Erhan, I.~Goodfellow, and
  R.~Fergus.
\newblock Intriguing properties of neural networks.
\newblock {\em arXiv preprint arXiv:1312.6199}, 2013.

\bibitem{tsogkas2012learning}
S.~Tsogkas and I.~Kokkinos.
\newblock Learning-based symmetry detection in natural images.
\newblock In {\em ECCV}, pages 41--54. Springer, 2012.

\bibitem{tzimiropoulos2014gauss}
G.~Tzimiropoulos and M.~Pantic.
\newblock Gauss-newton deformable part models for face alignment in-the-wild.
\newblock In {\em CVPR}, 2014.

\bibitem{fwang2013pose}
F.~Wang and Y.~Li.
\newblock Beyond physical connections: Tree models in human pose estimation.
\newblock In {\em CVPR}, 2013.

\bibitem{wang2007modeling}
P.~Wang, Q.~Ji, and J.~L. Wayman.
\newblock Modeling and predicting face recognition system performance based on
  analysis of similarity scores.
\newblock {\em TPAMI}, 29(4):665--670, 2007.

\bibitem{xiong2013supervised}
X.~Xiong and F.~De~la Torre.
\newblock Supervised descent method and its applications to face alignment.
\newblock In {\em CVPR}, 2013.

\bibitem{xiongSDMarxiv}
X.~Xiong and F.~De~la Torre.
\newblock Supervised descent method for solving nonlinear least squares
  problems in computer vision.
\newblock {\em arXiv:1405.0601}, 2014.

\bibitem{yang2013articulated}
Y.~Yang and D.~Ramanan.
\newblock Articulated human detection with flexible mixtures of parts.
\newblock {\em T-PAMI}, 35(12):2878--2890, 2013.

\bibitem{cfaneccv2014}
J.~Zhang, S.~Shan, M.~Kan, and X.~Chen.
\newblock Coarse-to-fine auto-encoder networks (cfan) for real-time face
  alignment.
\newblock In {\em ECCV}, 2014.

\bibitem{zhangpredicting}
P.~Zhang, J.~Wang, A.~Farhadi, M.~Hebert, and D.~Parikh.
\newblock Predicting failures of vision systems.
\newblock In {\em CVPR}, 2014.

\bibitem{devacvpr2012face}
X.~Zhu and D.~Ramanan.
\newblock {Face detection, pose estimation and landmark localization in the
  wild}.
\newblock In {\em CVPR}, 2012.

\end{thebibliography}
}

\end{document}